\documentclass{article}

\usepackage{microtype}
\usepackage{graphicx}
\usepackage{subfigure}
\usepackage{booktabs} 
\usepackage{amsmath}

\usepackage[hyphens]{url}
\usepackage{hyperref}
\usepackage{amsfonts}
\usepackage{amsmath}
\usepackage{cleveref}
\usepackage{mathtools}
\DeclareMathOperator*{\argmax}{argmax} 

\usepackage[accepted]{icml2020}

\icmltitlerunning{Adversarial Detection and Correction by Matching Prediction Distributions}

\newcounter{daggerfootnote}
\newcommand*{\daggerfootnote}[1]{%
    \setcounter{daggerfootnote}{\value{footnote}}%
    \renewcommand*{\thefootnote}{\fnsymbol{footnote}}%
    \footnote[2]{#1}%
    \setcounter{footnote}{\value{daggerfootnote}}%
    \renewcommand*{\thefootnote}{\arabic{footnote}}%
    }

\begin{document}

\twocolumn[
\icmltitle{Adversarial Detection and Correction by Matching Prediction Distributions}



\icmlsetsymbol{equal}{*}

\begin{icmlauthorlist}
    \icmlauthor{Giovanni Vacanti}{equal,seldon}
    \icmlauthor{Arnaud Van Looveren}{equal,seldon}
\end{icmlauthorlist}

\icmlaffiliation{seldon}{Seldon Technologies Ltd, London, United Kingdom}

\icmlcorrespondingauthor{Giovanni Vacanti}{gv@seldon.io}
\icmlcorrespondingauthor{Arnaud Van Looveren}{avl@seldon.io}

\icmlkeywords{Machine Learning, ICML}

\vskip 0.3in
]



\printAffiliationsAndNotice{\icmlEqualContribution} 

\begin{abstract}
We present a novel adversarial detection and correction method for machine learning classifiers. The detector consists of an autoencoder trained with a custom loss function based on the Kullback-Leibler divergence between the classifier predictions on the original and reconstructed instances. The method is unsupervised, easy to train and does not require any knowledge about the underlying attack. The detector almost completely neutralises powerful attacks like Carlini-Wagner or SLIDE on MNIST and Fashion-MNIST, and remains very effective on CIFAR-10 when the attack is granted full access to the classification model but not the defence. We show that our method is still able to detect the adversarial examples in the case of a white-box attack where the attacker has full knowledge of both the model and the defence and investigate the robustness of the attack. The method is very flexible and can also be used to detect common data corruptions and perturbations which negatively impact the model performance. We illustrate this capability on the CIFAR-10-C dataset.\daggerfootnote{An open source implementation of the algorithm can be found at \url{https://github.com/SeldonIO/alibi-detect}.}
\end{abstract}

\section{Introduction}

Adversarial examples \cite{szegedy2013intriguing} are instances which are carefully crafted by applying small perturbations to the original data with the goal to trick the machine learning classifier and change the predicted class. As a result, the classification model makes erroneous predictions which poses severe security issues for the deployment of machine learning systems in the real world. Achieving an acceptable level of security against adversarial attacks is a milestone that must be reached in order to trust and act on the predictions of safety-critical machine learning systems at scale. The issue regards many emerging technologies in which the use of machine learning models is prominent. For example, an attacker could craft adversarial images in order to induce an autonomous driving system to interpret a \emph{STOP} sign as a \emph{RIGHT OF WAY} sign and compromise the safety of the autonomous vehicle.

Given the crucial importance of the subject, a number of proposals claiming valid defence methods against adversarial attacks have been put forth in recent years (see \Cref{sec:related} for more details). Some of these proposals have obtained promising results on basic benchmark datasets for \emph{grey-box} attacks, i.e. attacks where the attacker has full knowledge of the model but not of the defence system. However, even in the grey-box scenario most of these approaches usually fail to generalise to more complex datasets or they are impractical. Moreover, effective and practical defences against \emph{white-box} attacks, i.e. attacks where the attacker has full knowledge of the model \emph{and} the defence mechanism, are still out of reach.

We argue that autoencoders trained with loss functions based on a distance metric between the input data and the reconstructed instances by the autoencoder network are flawed for the task of adversarial detection since they do not take the goal of the attacker into account. The attack applies near imperceptible perturbations to the input which change the class predicted by the classifier. Since the impact of the attack is most prominent in the model output space, this is where the defence should focus on during training. We propose a novel method for adversarial detection and correction based on an autoencoder network with a model dependent loss function designed to match the prediction probability distributions of the original and reconstructed instances. The output of the autoencoder can be seen as a \emph{symmetric example} since it is crafted to mimic the prediction distribution produced by the classifier on the original instance and does not contain the adversarial artefact anymore. We also define an \emph{adversarial score} based on the mismatch between the prediction distributions of the classifier on an instance and its symmetric counterpart. This score is highly effective to detect both grey-box and white-box attacks. The defence mechanism is trained in an unsupervised fashion and does not require any knowledge about the underlying attack. Because the method is designed to extract knowledge from the classifier's output probability distribution, it bears some resemblance to \emph{defensive distillation} \cite{defensivedistillation}.

Besides detecting malicious adversarial attacks, the adversarial score also proves to be an effective measure for more common data corruptions and perturbations which degrade the machine learning model's performance. Our method is in principle applicable to any machine learning classifier vulnerable to adversarial attacks, regardless of the modality of the data.

In the following, \Cref{sec:related} gives a brief summary of current developments in the field of adversarial defence. In \Cref{sec:method} we describe our method in more detail while \Cref{sec:experiments} discusses the results of our experiments. We validate our method against a variety of state-of-the-art grey-box and white-box attacks on the MNIST \cite{lecun2010mnist}, Fashion-MNIST \cite{fashionmnist} and CIFAR-10 \cite{cifar10} datasets. We also evaluate our method as a data drift detector on the CIFAR-10-C dataset \cite{HendrycksD19}.

\section{Related Work}\label{sec:related}
\subsection{Adversarial attacks}

Since the discovery of adversarial examples \cite{szegedy2013intriguing}, a variety of methods have been proposed to generate such instances via adversarial attacks. The aim of the attack is to craft an instance $x_{\text{adv}}$ that changes the predicted class $c$ of the machine learning classifier $M$ without noticeably altering the original instance $x$. In other words, the attack tries to find the smallest perturbation $\delta$ such that the model predicts different classes for $x$ and $x + \delta$. 

If the attack is \emph{targeted}, the classifier prediction on $x + \delta$ is restricted to a predetermined class $c$. For an \emph{untargeted} attack, any class apart from the one predicted on $x$ is sufficient for the attack to be successful. From here on we only consider untargeted attacks as they are less constricted and easier to succeed. Formally, the attack tries to solve the following optimisation problem:

\begin{equation}\label{eq:adv_minin}
\min_{\delta}||\delta||_p \quad s.t.\quad C(x) \neq C(x + \delta) 
\end{equation}

where $||\cdot||_p$ is the $\ell_p$ norm, $C(\cdot) = \argmax M(\cdot)$ and $M(\cdot)$ represents the prediction probability vector of the classifier.

We validate our adversarial defence on three different attacks: Carlini-Wagner \cite{carlini2016evaluating}, SLIDE \cite{tramr2019adversarial} and the Fast Gradient Sign Method (FGSM) \cite{fgsm}. The Carlini-Wagner (C\&W) and SLIDE attacks are very powerful and able to reduce the accuracy of machine learning classifiers on popular datasets such as CIFAR-10 to nearly $0$\% while keeping the adversarial instances visually indistinguishable from the original ones. FGSM on the other hand is a fast but less powerful attack, resulting in more obvious adversarial instances. Although many other attack methods like EAD \cite{eadattack}, DeepFool \cite{deepfool} or Iterative FGSM \cite{iterativefgsm} exist, the scope covered by C\&W, SLIDE and FGSM is sufficient to validate our defence.

\paragraph{Carlini-Wagner}
We use the $\ell_2$ version of the C\&W attack. The $\ell_2$-C\&W attack approximates the minimisation problem given in \Cref{eq:adv_minin} as
\begin{equation}
\min_{\delta} ||\delta||_2 + c \cdot f(x + \delta) \quad s.t. \quad x + \delta \in [0,1]^n    
\end{equation}
where $f(x + \delta)$ is a custom loss designed to be negative if and only if the class predicted by the model for $x + \delta$ is equal to the target class and where $||\cdot||_2$ denotes the $\ell_2$ norm. 

\paragraph{SLIDE}
The SLIDE attack is an iterative attack based on the $\ell_1$ norm. Given the loss function of the classifier $L(\theta, x, y),$ at each iteration the gradients ${\bf g}$ with respect to the input are calculated. The unit vector ${\bf e}$ determines the direction of the perturbation $\delta$ and the components of ${\bf e}$ are updated according to
\begin{equation}
    e_i = \text{sign}(g_i)  \quad \text{if} \quad |g_i| > P_q(|{\bf g}|) \quad \text{else} \quad 0 
\end{equation}
where $P_q(|{\bf g}|)$ represent the $q$-th percentile of the gradients' components. 
The perturbation $\delta$ is then updated as $\delta \leftarrow \delta + \lambda \cdot {\bf e} / ||{\bf e}||_2$ where $\lambda$ is the step size of the attack. 
\paragraph{FGSM} 
The Fast Gradient Sign Method is designed to craft a perturbation $\delta$ in the direction of the gradients of the model's loss function with respect to the input $x$ according to 
\begin{equation}
    x + \delta = x + \epsilon \cdot \text{sign}(\Vec{\nabla}_{x}L(\theta, x, y)),
\end{equation}
where $\epsilon$ is a small parameter fixing the size of the perturbation and $L(\theta, x, y)$ is the loss function of the classifier. 

\subsection{Adversarial Defences}

Different defence mechanisms have been developed to deal with the negative impact of adversarial attacks on the classifier's performance. Adversarial training augments the training data with adversarial instances to increase the model robustness \cite{szegedy2013intriguing, Goodfellow2015}. Adversarial training tailored to a specific attack type can however leave the model vulnerable to other perturbation types \cite{tramr2019adversarial}.

A second approach attempts to remove the adversarial artefacts from the example and feed the \emph{purified} instance to the machine learning classifier. \emph{Defense-GAN} \cite{Samangouei2020} uses a Wasserstein GAN \cite{arjovsky2017wasserstein} which is trained on the original data. The difference between the GAN’s \cite{goodfellowgan} generator output $G(z)$ and the adversarial instance $x_{\text{adv}}$ is minimised with respect to $z$. The generated instance $G(z^*)$ is then fed to the classifier. Defense-GAN comes with a few drawbacks. GAN training can be notoriously unstable and suffer from issues like mode collapse which would reduce the effectiveness of the defence. The method also needs to apply $L$ gradient descent optimisation steps with $R$ random restarts at inference time, making it computationally more expensive. \emph{MagNet} \cite{meng2017magnet} uses one or more autoencoder-based detectors and reformers to respectively flag adversarial instances and transform the input data before feeding it to the classifier. The autoencoders are trained with the mean squared error (MSE) reconstruction loss, which is suboptimal for adversarial detection as it focuses on reconstructing the input without taking the decision boundaries into account. Other defences using autoencoders either require knowledge about the attacks like \cite{li2019purifying, defensevae} or class labels \cite{puvae}. \emph{PixelDefend} \cite{pixeldefend} uses a generative model to purify the adversarial instance.

A third approach, \emph{defensive distillation} \cite{defensivedistillation}, utilises model distillation \cite{hinton2015distilling} as a defence mechanism. A second classification model is trained by minimising the cross entropy between the class probabilities of the original classifier and the predictions of the distilled model. Defensive distillation reduces the impact of gradients used in crafting adversarial instances and increases the number of features that need to be changed. Although our method uses an autoencoder to mitigate the adversarial attack, it can also relate to defensive distillation since we optimise for the K-L divergence between the model predictions on the original and reconstructed instances.

\section{Method}\label{sec:method}
\subsection{Threat Model}

The threat model describes the capabilities and knowledge of the attacker. These assumptions are crucial to evaluate the defence mechanism and allow like-for-like comparisons.

The attack is only allowed to change the \emph{input features} $x$ by a small perturbation $\delta$ such that the predicted class $C(x + \delta)$ is different from $C(x)$. Since we assume that the attack is untargeted, $C(x + \delta)$ is not restricted to a predefined class $c$. The attack is however not allowed to modify the weights or architecture of the machine learning model. A most important part of the threat model is the knowledge of the attack about both the model and the defence. We consider three main categories:

\paragraph{Black-box attacks}
This includes all attack types where the attacker only has access to the input and output of the classifier under attack. The output can either be the predicted class or the output probability distribution over all the classes.

\paragraph{Grey-box attacks}
The attacker has full knowledge about the classification model but not the defence mechanism. This includes information about the model's architecture, weights, loss function and gradients.

\paragraph{White-box attacks}
On top of full knowledge about the classifier, the attack also has complete access to the internals of the defence mechanism. This includes the logic of the defence, loss function as well as model weights and gradients for a differentiable defence system. Security against white-box attacks implies security against all the less powerful grey-box and black-box attacks.

We validate the strength of our proposed defence mechanism for a variety of grey-box and white-box attacks on the MNIST, Fashion-MNIST and CIFAR-10 datasets.

\subsection{Defence Mechanism}\label{sec:Defence Mechanism}

\begin{figure}[!t]
\vskip 0.2in
\begin{center}
\centerline{\includegraphics[width=\columnwidth]{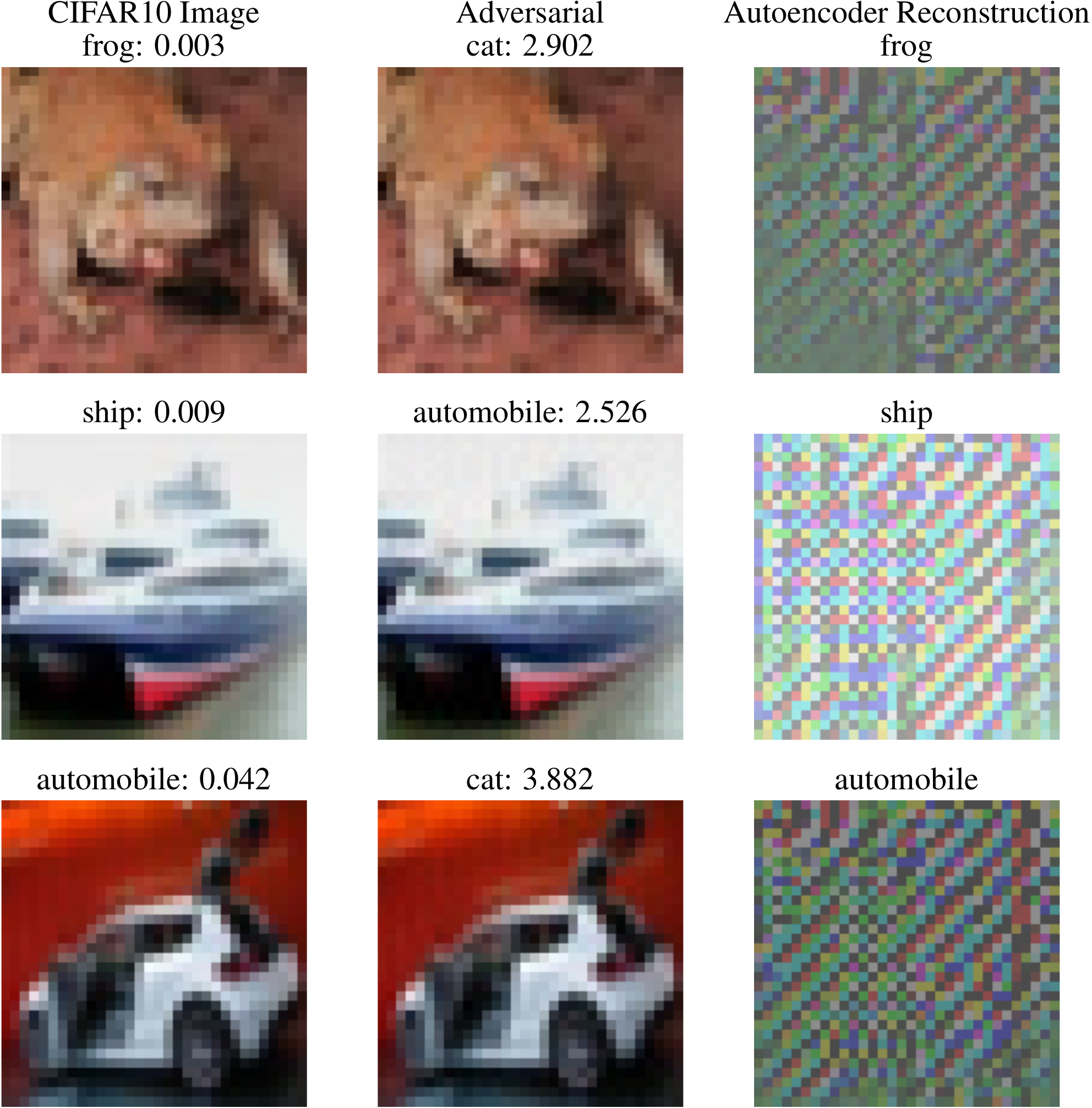}}
\caption{Adversarial detection and correction examples on CIFAR-10 after a C\&W attack. The first two columns show respectively the original and adversarial instances with the class predictions and adversarial scores. The last column visualises the reconstructed instance of the adversarial image by the autoencoder with the corrected prediction.}
\label{fig:fig_cifar10_example}
\end{center}
\vskip -0.2in
\end{figure}

Our novel approach is based on an autoencoder network. An autoencoder $AE$ consists of an encoder which maps vectors $x$ in the input space $\mathbb{R}^n$ to vectors $z$ in a \emph{latent space} $\mathbb{R}^d$ with $d < n$, and a decoder which maps $z$ back to vectors $x' = AE(x)$ in $\mathbb{R}^n$. The encoder and decoder are jointly trained to approximate an input transformation $T: \mathcal{X} \rightarrow \mathcal{X},$ which is defined by the optimisation objective, or loss function.

There have been multiple attempts to employ conventionally trained autoencoders for adversarial example detection \cite{meng2017magnet, puvae, li2019purifying}. Usually, autoencoders are trained to find a transformation $T$ that reconstructs the input instance ${x}$ as accurately as possible with loss functions that are suited to capture the similarities between $x$ and $x^\prime$ such as the reconstruction error $||x - x'||_2^2$. However, these types of loss functions suffer from a fundamental flaw for the task of adversarial detection and correction. In essence, the attack tries to introduce a minimal perturbation $\delta$ in the input space while maximising the impact of $\delta$ on the model output space to ensure $C(x) \neq C(x + \delta)$. If the autoencoder $AE_{\theta}^{\text{MSE}}$ is trained with a reconstruction error loss, $x'$ will lie very close to $x$ and will be sensitive to the same adversarial perturbation $\delta$ crafted around $x$. There is no guarantee that the transformation $AE_{\theta}^{\text{MSE}}(x + \delta)$ is able to remove the adversarial perturbation from the input since the autoencoder's objective is only to reconstruct $x + \delta$ as truthful as possible in the input space.

The novelty of our proposal relies on the use of a model-dependent loss function based on a distance metric in the output space of the model to train the autoencoder network. Given a model $M$ we optimise the weights $\theta$ of an auto-encoder $AE_{\theta}$ using the following objective function:
\begin{equation}\label{eq:lossAE}
    \min_{\theta} D_{\text{KL}}(M(x) || M(AE_{\theta}^{\text{KL}}(x)))
\end{equation}
where $D_{\text{KL}}(\cdot||\cdot)$ denotes the K-L divergence and $M(\cdot)$ represents the prediction probability vector of the classifier. Training of the autoencoder is unsupervised since we only need access to the model prediction probabilities and the normal training instances. The classifier weights are frozen during training. Note that the fundamental difference between our approach and other defence systems based on autoencoders relies on the fact that the minimisation objective is suited to capture similarities between instances in the output space of the model rather than in the input feature space.

Without the presence of a reconstruction loss term like $||x - x'||_2^2$, $x'$ simply tries to make sure that the prediction probabilities $M(x')$ and $M(x)$ match without caring about the proximity of $x'$ to $x$. As a result, $x'$ is allowed to live in different areas of the input feature space than $x$ with different decision boundary shapes with respect to the model $M$. The carefully crafted adversarial perturbation $\delta$ which is effective around $x$ does not transfer to the new location of $x'$ in the feature space, and the attack is therefore neutralised. This effect is visualised by \Cref{fig:fig_cifar10_example}. The adversarial instance is close to the original image $x$ in the pixel space but the reconstruction of the adversarial attack by the autoencoder $x'_{\text{adv}} = AE_{\theta}^{\text{KL}}(x_{\text{adv}})$ lives in a different region of the input space than $x_{\text{adv}}$ and looks like noise at first glance. $x'_{\text{adv}}$ does however not contain the adversarial artefacts anymore and the model prediction $C(x'_{\text{adv}})$ returns the corrected class.

The adversarial instances can also be detected via the adversarial score $S_{\text{adv}}$:
\begin{equation}\label{eq:adv_score}
    S_{\text{adv}}(x) = D(M(x) || M(AE_{\theta}^{\text{KL}}(x)))
\end{equation}
where $D(\cdot||\cdot)$ is again a distance metric like the K-L divergence. $S_{\text{adv}}$ will assume high values for adversarial examples given the probability distribution difference between predictions on the adversarial and reconstructed instance, making it a very effective measure for adversarial detection.

\begin{figure}[!t]
\vskip 0.2in
\begin{center}
\centerline{\includegraphics[width=\columnwidth]{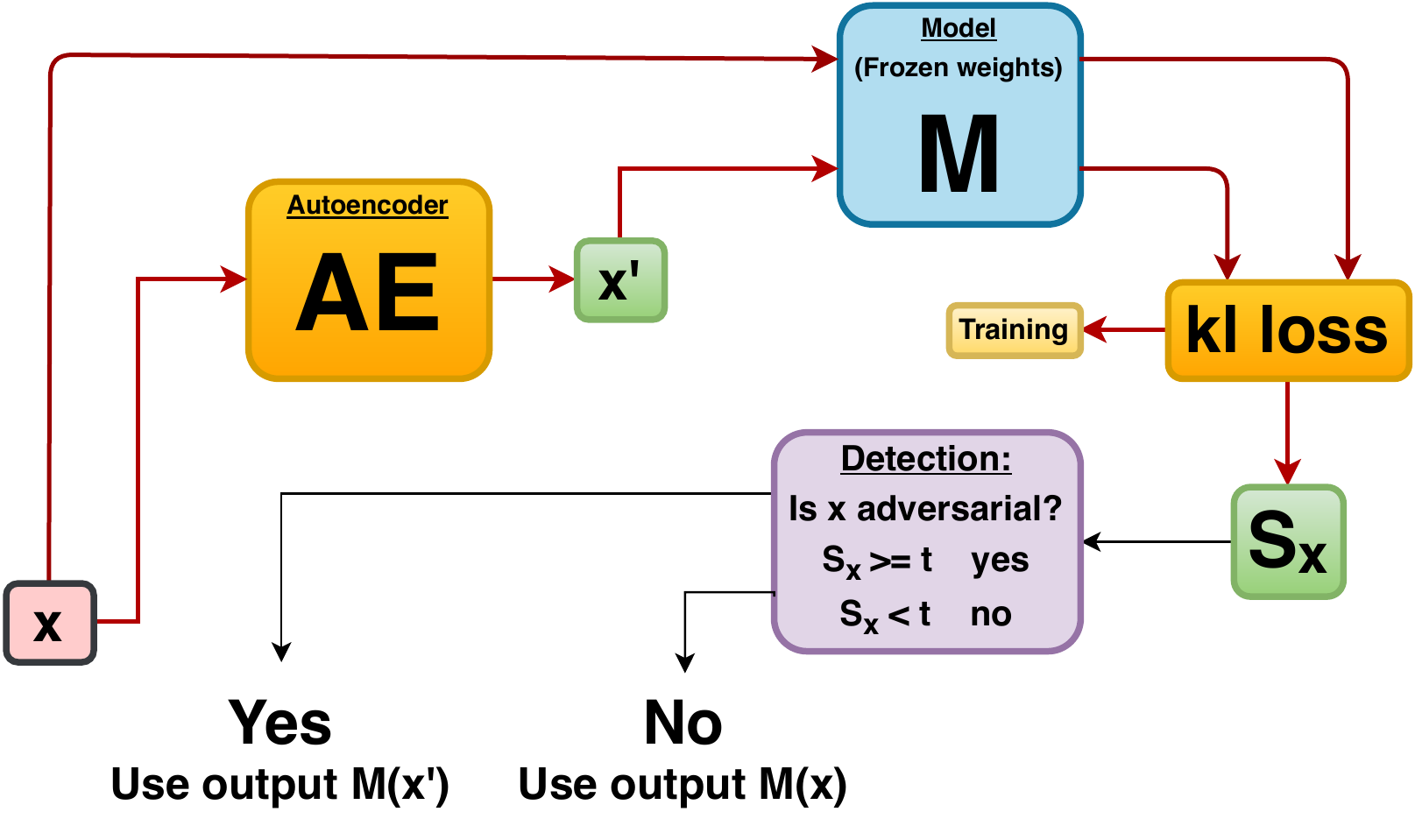}}
\caption{An input instance ${x}$ is transformed into $x^\prime$ by the autoencoder $AE.$ The K-L divergence between the output distributions $M({x})$ and $M({x^\prime})$ is calculated and used as loss function for training $AE$ and as the adversarial signal $S_x$ at inference time. Based on some appropriate threshold $t$, $S_x$ is used to flag adversarial instances. If an instance is flagged, the correct prediction is retrieved through the transformed instance $x^\prime.$ Note that the model's weights are frozen during training.}
\label{fig:method_diagram}
\end{center}
\vskip -0.2in
\end{figure}

Both detection and correction mechanisms can be combined in a simple yet effective adversarial defence system, illustrated in \Cref{fig:method_diagram}. First the adversarial score $S_{\text{adv}}(x)$ is computed. If the score exceeds a predefined threshold $t$ the instance is flagged as an adversarial example $x_{\text{adv}}$. Similar to MagNet the threshold $t$ can be set using only normal data by limiting the false positive rate to a small fraction $\epsilon_{\text{FPR}}$. The adversarial instance is fed to the autoencoder which computes the transformation $x'_{\text{adv}} = AE_{\theta}^{\text{KL}}(x_{\text{adv}})$. The classifier $M$ finally makes a prediction on $x'_{\text{adv}}$. If the adversarial score is below the threshold $t$, the model makes a prediction on the original instance $x$.

The method is also well suited for drift detection, i.e. for the detection of corrupted or perturbed instances that are not necessarily adversarial by nature but degrade the model performance.

\subsection{Method Extensions}

The performance of the correction mechanism can be improved by extending the training methodology to one of the hidden layers. We extract a flattened feature map $F$ from the hidden layer, feed it into a linear layer and apply the softmax function:

\begin{equation}\label{eq:hl}
    y_{\psi}(x) = \text{softmax}(W_{\psi} F(x) + b).
\end{equation}

The autoencoder is then trained by optimising

\begin{equation}\label{eq:loss_hl}
\begin{split}
    \min_{\theta, \psi} D_{\text{KL}}(M(x) || M(AE_{\theta}^{\text{KL}}(x)) + \\
    \lambda D_{\text{KL}}(y_{\psi}(x) || y_{\psi}(AE_{\theta}^{\text{KL}}(x))).
\end{split}
\end{equation}

During training of $AE_{\theta}^{\text{KL}}$, the K-L divergence between the model predictions on $x$ and $x'$ is minimised. If the entropy from the output of the model's softmax layer is high, it becomes harder for the autoencoder to learn clear decision boundaries. In this case, it can be beneficial to sharpen the model's prediction probabilities through temperature scaling:

\begin{equation}\label{eq:temperature_scaling}
    M(x)_T = \frac{M(x)^\frac{1}{T}}{\sum_{j} M(x)_j^\frac{1}{T}}.
\end{equation}

The loss to minimise becomes $D_{\text{KL}}(M(x)_T || M(AE_{\theta}^{\text{KL}}(x)))$. The temperature $T$ itself can be tuned on a validation set.

\section{Experiments}\label{sec:experiments}

\subsection{Experimental Setup}

The adversarial attack experiments are conducted on the MNIST, Fashion-MNIST and CIFAR-10 datasets. The autoencoder architecture is similar across the datasets and consists of 3 convolutional layers in both the encoder and decoder. The MNIST and Fashion-MNIST classifiers are also similar and reach test set accuracies of respectively $99.28$\% and $93.62$\%. For CIFAR-10, we train both a stronger ResNet-56 \cite{resnet} model up to $93.15$\% accuracy and a weaker model which achieves $80.24$\% accuracy on the test set. More details about the exact architecture and training procedure of the different models can be found in the appendix.

The defence mechanism is tested against Carlini-Wagner (C\&W), SLIDE and FGSM attacks with varying perturbation strength $\epsilon$. The attack hyperparameters and examples of adversarial instances for each attack can be found in the appendix. The attacks are generated using the open source Foolbox library \cite{rauber2017foolbox}.

We consider two settings under which the attacks take place: grey-box and white-box. Grey-box attacks have full knowledge of the classification model $M$ but not the adversarial defence. White-box attacks on the other hand assume full knowledge of both the model and defence and can propagate gradients through both. As a result, white-box attacks try to fool $C(AE_{\theta}^{\text{KL}}(x))$.

\subsection{Grey-Box Attacks}

\begin{figure}[!t]
\vskip 0.2in
\begin{center}
\centerline{\includegraphics[width=\columnwidth]{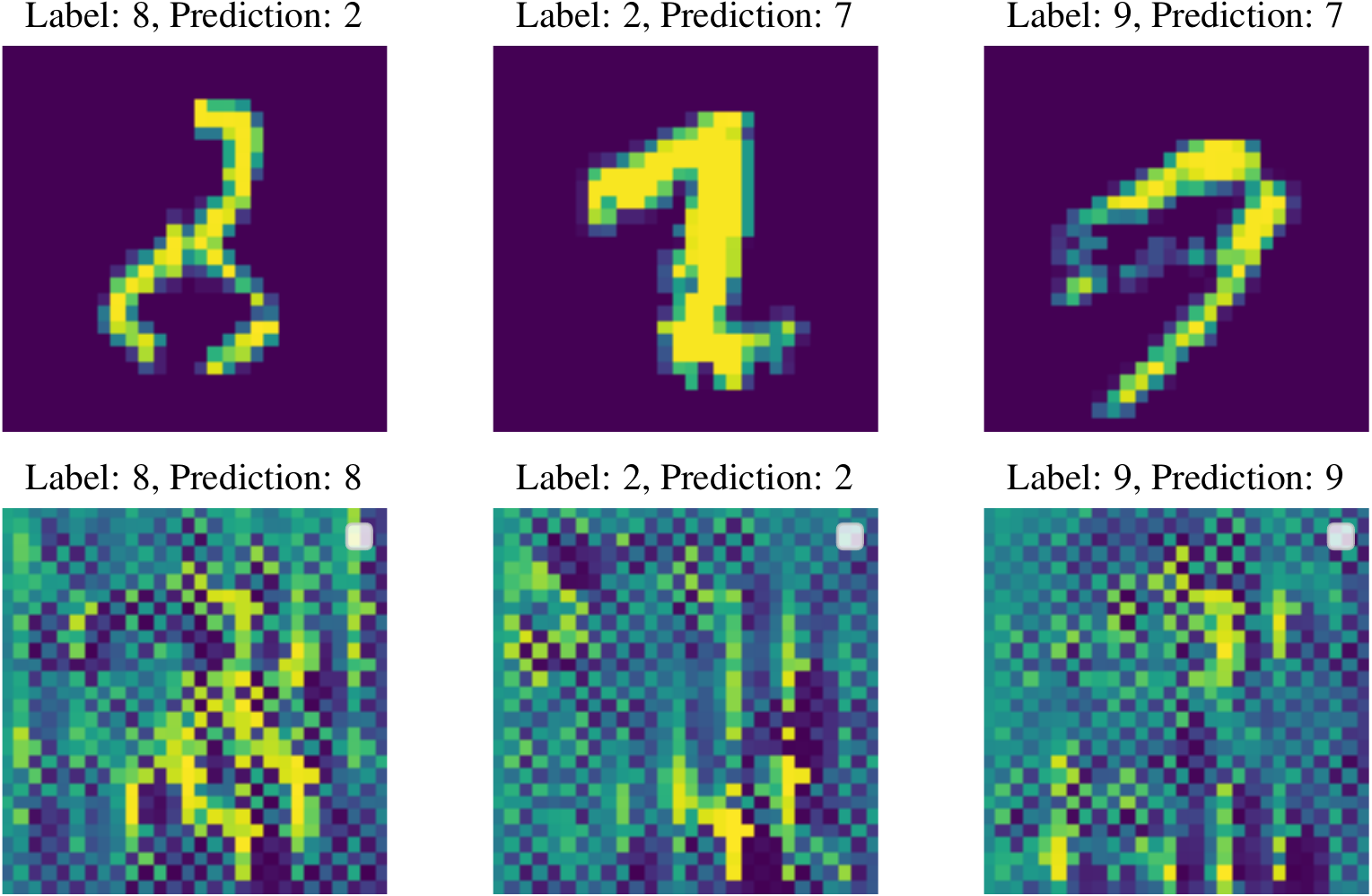}}
\caption{The reconstructed images by the adversarial autoencoder in the bottom row correct classifier mistakes on MNIST.}
\label{fig:fig_corr_mnist}
\end{center}
\vskip -0.2in
\end{figure}

\begin{table}[!b]
    \centering
    \caption{Test set accuracy for the MNIST classifier on both the original and adversarial instances with and without the defence. $AE^{\text{MSE}}$ and $AE^{\text{KL}}$ are the defence mechanisms trained with respectively the MSE and $D_{\text{KL}}$ loss functions.}
    \vskip 0.15in
    \resizebox{\columnwidth}{!}{
    \begin{tabular}{@{}ccccc@{}}
    \toprule
    \textbf{Attack} & \textbf{No Attack} & \textbf{No Defence} & $\mathbf{AE^{\text{MSE}}}$ & $\mathbf{AE^{\text{KL}}}$\\ \midrule
    CW & $0.9928$ & $0.0058$ & $0.9887$ & $\mathbf{0.9928}$ \\
    SLIDE & $0.9928$ & $0.0001$ & $0.9891$ & $\mathbf{0.9923}$ \\
    FGSM, $\epsilon=0.1$ & $0.9928$ & $0.9109$ & $0.9901$ & $\mathbf{0.9928}$ \\
    FGSM, $\epsilon=0.2$ & $0.9928$ & $0.5009$ & $0.9821$ & $\mathbf{0.9895}$ \\
    FGSM, $\epsilon=0.3$ & $0.9928$ & $0.1711$ & $0.9610$ & $\mathbf{0.9819}$ \\ \bottomrule
    \end{tabular}
    }
    \label{tb:table_mnist}
\end{table}

\begin{table}[!b]
    \centering
    \caption{Test set accuracy for the Fashion-MNIST classifier on both the original and adversarial instances with and without the defence. $AE^{\text{MSE}}$ and $AE^{\text{KL}}$ are the defence mechanisms trained with respectively the MSE and $D_{\text{KL}}$ loss functions.}
    \vskip 0.15in
    \resizebox{\columnwidth}{!}{%
    \begin{tabular}{@{}cccccc@{}}
    \toprule
    \textbf{Attack} & \textbf{No Attack} & \textbf{No Defence} & $\mathbf{AE^{\text{MSE}}}$ & $\mathbf{AE^{\text{KL}}}$ \\ \midrule
    CW & $0.9362$ & $0.1050$ & $0.9173$ & $\mathbf{0.9211}$ \\
    SLIDE & $0.9362$ & $0.0000$ & $0.9211$ & $\mathbf{0.9247}$ \\
    FGSM, $\epsilon=0.1$ & $0.9362$ & $0.1305$ & $0.9132$ & $\mathbf{0.9169}$ & \\
    FGSM, $\epsilon=0.2$ & $0.9362$ & $0.0706$ & $0.9078$  & $\mathbf{0.9111}$ & \\
    FGSM, $\epsilon=0.3$ & $0.9362$ & $0.0529$ & $0.9007$ & $\mathbf{0.9055}$ & \\ \bottomrule
    \end{tabular}%
    }
    \label{tb:table_fashion_mnist}
\end{table}

Mitigating adversarial attacks on classification tasks consists of two steps: detection and correction. \Cref{tb:table_mnist} to \Cref{tb:table_cifar10_strong} highlight the consistently strong performance of the correction mechanism across the different datasets for various attack types. On MNIST, strong attacks like C\&W and SLIDE which reduce the model accuracy to almost $0$\% are corrected by the detector and the attack is neutralised, nearly recovering the original accuracy of $99.28$\%. It is also remarkable that when we evaluate the accuracy of the classifier predictions $C(AE_{\theta}^{\text{KL}}(x))$ where $x$ is the original test set, the accuracy equals $99.44$\%, surpassing the performance of $C(x)$. \Cref{fig:fig_corr_mnist} shows a few examples of instances ${x}$ that are corrected by $AE_{\theta}^{\text{KL}}$, as well as their reconstruction $x'$. The corrected instances are outliers in the pixel space for which the autoencoder manages to capture the decision boundary. The correction accuracy drops slightly from $99.28$\% to $98.19$\% for FGSM attacks when increasing the perturbation strength $\epsilon$ from $0.1$ to $0.3$. Higher values of $\epsilon$ lead to noisier adversarial instances which are easy to spot with the naked eye and result in higher adversarial scores. The results on Fashion-MNIST follow the same narrative. The adversarial correction mechanism largely restores the model accuracy after powerful attacks which can reduce the accuracy without the defence up to $0$\%.

\begin{table}[!b]
    \centering
    \caption{CIFAR-10 test set accuracy for a simple CNN classifier on both the original and adversarial instances with and without the defence. $AE^{\text{MSE}}$ and $AE^{\text{KL}}$ are the defence mechanisms trained with respectively the MSE and $D_{\text{KL}}$ loss functions. $AE^{\text{KL, T}}$ includes temperature scaling and $AE^{\text{KL, HL}}$ extends the methodology to one of the hidden layers.}
    \vskip 0.15in
    \resizebox{\columnwidth}{!}{%
    \begin{tabular}{@{}ccccccc@{}}
    \toprule
    \textbf{Attack} & \textbf{No Attack} & \textbf{No Defence} & $\mathbf{AE^{\text{MSE}}}$ & $\mathbf{AE^{\text{KL}}}$  & $\mathbf{AE^{\text{KL, T}}}$ & $\mathbf{AE^{\text{KL, HL}}}$\\ \midrule
    CW & $0.8024$ & $0.0001$ & $0.6022$ & $0.7551$ & $0.7669$ & $\mathbf{0.7688}$ \\
    SLIDE & $0.8024$ & $0.0208$ & $0.6136$ & $0.7704$ & $0.7840$ & $\mathbf{0.7864}$ \\
    FGSM, $\epsilon=0.1$ & $0.8024$ & $0.0035$ & $0.5903$ & $0.7554$ & $0.7554$ & $\mathbf{0.7628}$ \\
    FGSM, $\epsilon=0.2$ & $0.8024$ & $0.0035$ & $0.5901$ & $0.7555$ & $0.7554$ & $\mathbf{0.7624}$ \\
    FGSM, $\epsilon=0.3$ & $0.8024$ & $0.0036$ & $0.5900$ & $0.7552$ & $0.7555$ & $\mathbf{0.7621}$ \\ \bottomrule
    \end{tabular}%
    }
    \label{tb:table_cifar10_weak}
\end{table}

\begin{table}[!b]
    \centering
    \caption{CIFAR-10 test set accuracy for a ResNet-56 classifier on both the original and adversarial instances with and without the defence. $AE^{\text{MSE}}$ and $AE^{\text{KL}}$ are the defence mechanisms trained with respectively the MSE and $D_{\text{KL}}$ loss functions. $AE^{\text{KL, T}}$ includes temperature scaling and $AE^{\text{KL, HL}}$ extends the methodology to one of the hidden layers.}
    \vskip 0.15in
    \resizebox{\columnwidth}{!}{%
    \begin{tabular}{@{}ccccccc@{}}
    \toprule
    \textbf{Attack} & \textbf{No Attack} & \textbf{No Defence} & $\mathbf{AE^{\text{MSE}}}$ & $\mathbf{AE^{\text{KL}}}$  & $\mathbf{AE^{\text{KL, T}}}$ & $\mathbf{AE^{\text{KL, HL}}}$\\ \midrule
    CW & $0.9315$ & $0.0000$ & $0.1650$ & $0.8048$ & $0.8141$ & $\mathbf{0.8153}$ \\
    SLIDE & $0.9315$ & $0.0000$ & $0.1659$ & $0.8159$ & $0.8265$ & $\mathbf{0.8360}$ \\
    FGSM, $\epsilon=0.1$ & $0.9315$ & $0.0140$ & $0.1615$ & $0.7875$ & $0.7882$ & $\mathbf{0.7957}$ \\
    FGSM, $\epsilon=0.2$ & $0.9315$ & $0.0008$ & $0.1615$ & $0.7781$ & $0.7760$ & $\mathbf{0.7908}$ \\
    FGSM, $\epsilon=0.3$ & $0.9315$ & $0.0000$ & $0.1615$ & $0.7772$ & $0.7752$ & $\mathbf{0.7899}$ \\ \bottomrule
    \end{tabular}%
    }
    \label{tb:table_cifar10_strong}
\end{table}

\begin{figure}[!ht]
\vskip 0.2in
\begin{center}
\centerline{\includegraphics[width=\columnwidth]{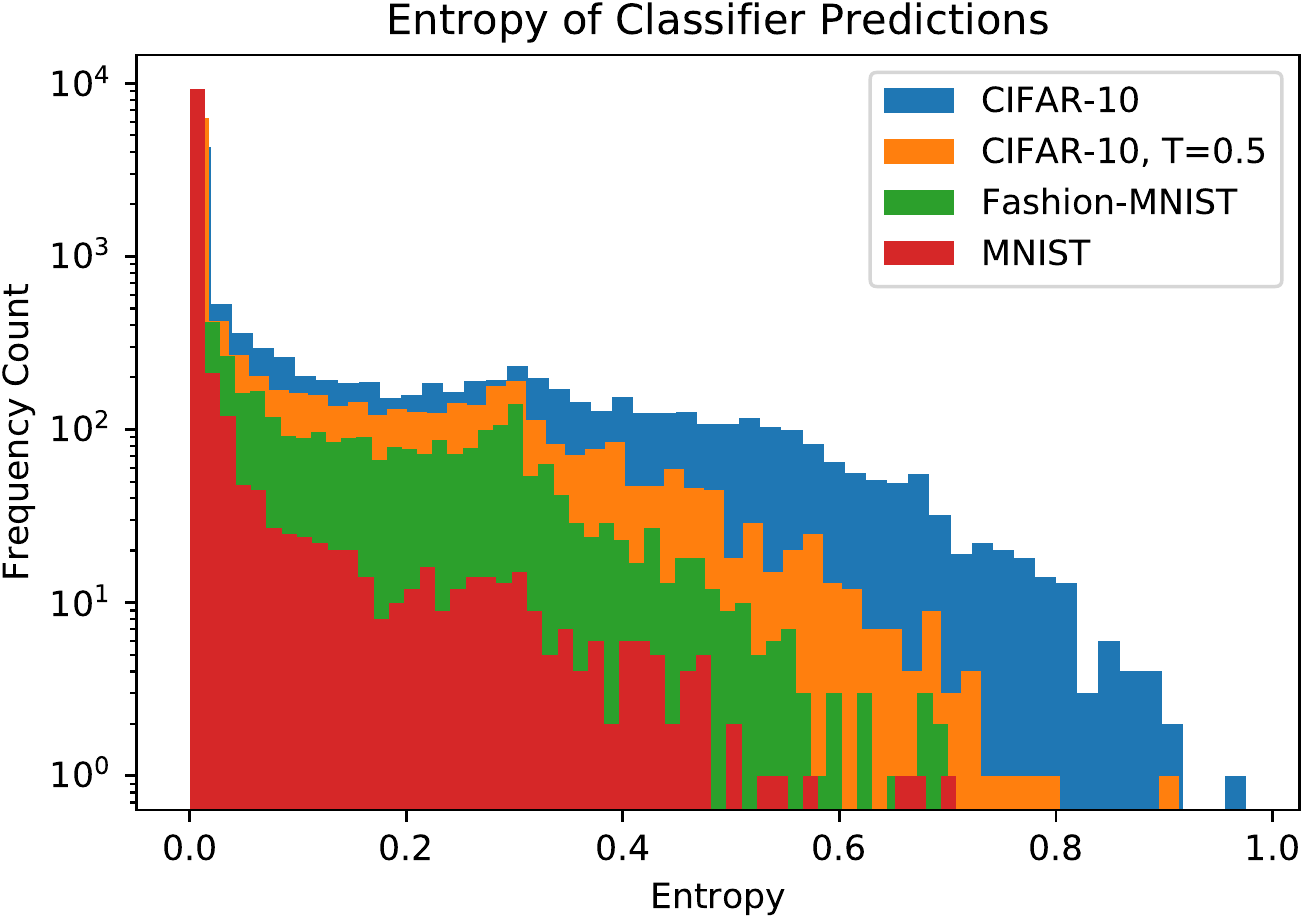}}
\caption{Entropy of the classifier predictions on the test set for MNIST, Fashion-MNIST and CIFAR-10.}
\label{fig:fig_entropy}
\end{center}
\vskip -0.2in
\end{figure}

The classification accuracy uplift on MNIST or Fashion-MNIST from training the autoencoder with $D_\text{KL}$ instead of the mean squared error between $x$ and $x'$ is limited from $0.3$\% to $0.4$\% for adversarial instances generated by C\&W or SLIDE. \Cref{tb:table_cifar10_weak} and \Cref{tb:table_cifar10_strong} however show that the autoencoder defence mechanism trained with the K-L divergence $AE_{\theta}^{D_{\text{KL}}}$ outperforms the MSE equivalent $AE_{\theta}^{\text{MSE}}$ by respectively over 15\% and almost 65\% on CIFAR-10 using the simple classifier and the ResNet-56 model. The performance difference is even more pronounced when we simplify the autoencoder architecture. An autoencoder with only one hidden dense layer and ReLU activation function \cite{hahnloserrelu} in the encoder and one hidden dense layer before the output layer in the decoder is still able to detect and correct adversarial attacks on the CIFAR-10 ResNet-56 model when trained with $D_\text{KL}(M(x) || M(AE_{\theta}^{\text{KL}}(x)))$. The correction accuracy for C\&W and SLIDE reaches $52.75$\% and $52.83$\% compared to around $10$\%, or similar to random predictions, if the same autoencoder is trained with the MSE loss. The exact architecture can be found in the appendix.

\begin{figure*}[t]
\centering
\includegraphics[width=0.95\textwidth]{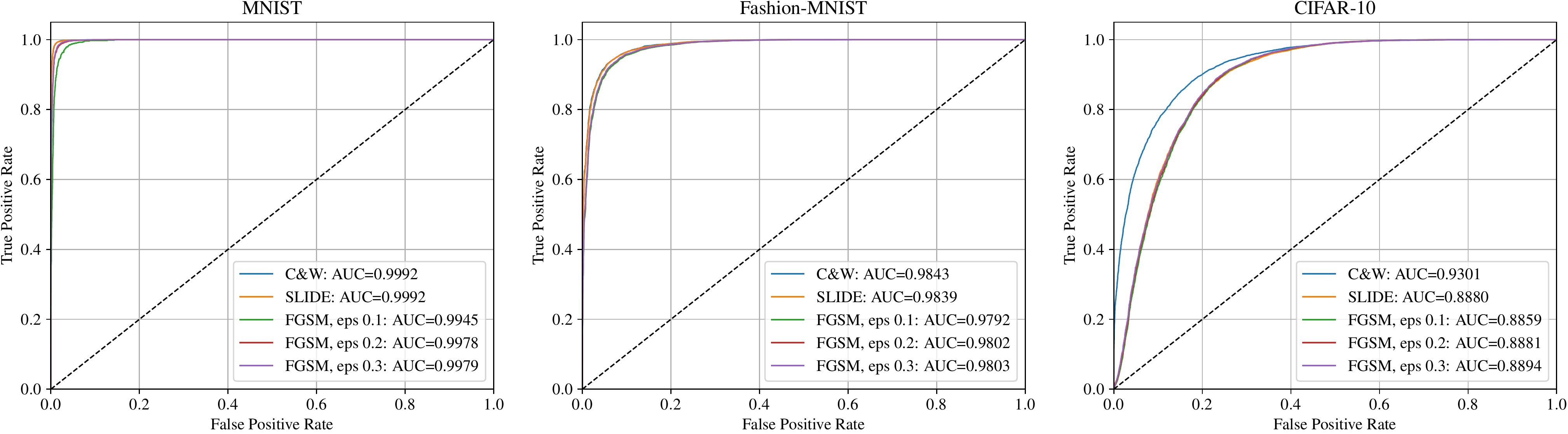}
\caption{ROC curves and AUC values for adversarial instance detection on MNIST, Fashion-MNIST and CIFAR-10 for C\&W, SLIDE and FGSM grey-box attacks. The curves and values are computed on the combined original and attacked test sets.}
\label{fig:fig_roc_all}
\end{figure*}

The ROC curves for the adversarial scores $S_\text{adv}$ and corresponding AUC values in \Cref{fig:fig_roc_all} highlight the effectiveness of the method for the different datasets. The AUC for the strong C\&W and SLIDE grey-box attacks are equal to $0.9992$ for MNIST and approximately $0.984$ for Fashion-MNIST. For the ResNet-56 classifier on CIFAR-10, the AUC is still robust at $0.9301$ for C\&W and $0.8880$ on SLIDE. As expected, increasing $\epsilon$ for the FGSM attacks results in slightly higher AUC values.

\begin{figure*}[t]
\centering
\includegraphics[width=0.95\textwidth]{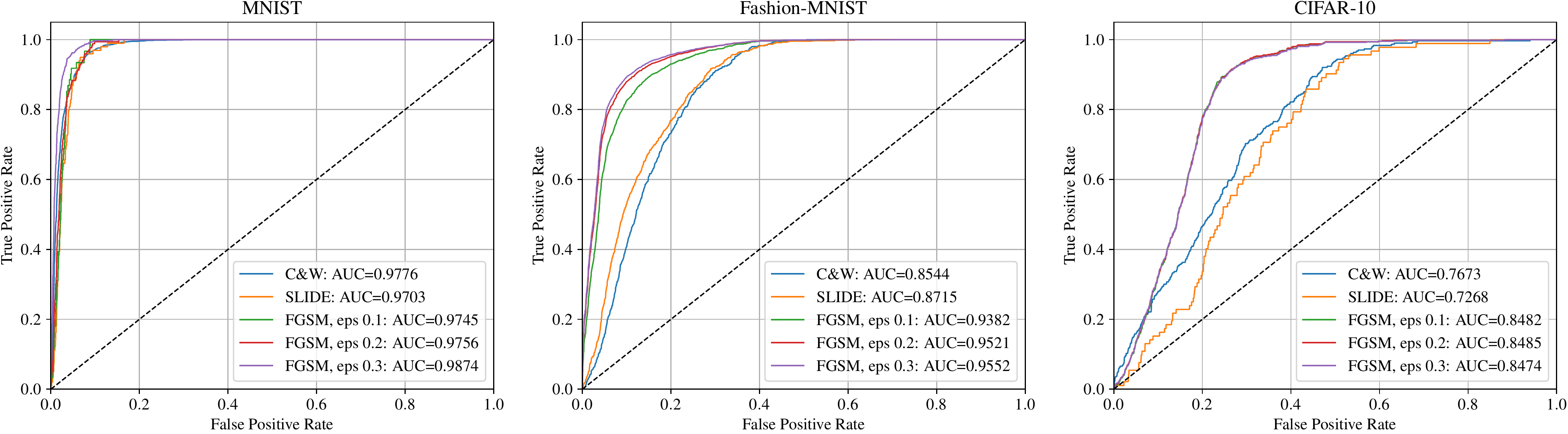}
\caption{ROC curves and AUC values for adversarial instance detection on MNIST, Fashion-MNIST and CIFAR-10 for C\&W, SLIDE and FGSM white-box attacks. The curves and values are computed on the combined original and attacked test sets.}
\label{fig:fig_roc_wb}
\end{figure*}

\Cref{fig:fig_entropy} shows that the classifier's entropy on CIFAR-10 is higher than on MNIST or Fashion-MNIST. As a result, we apply temperature scaling on $M(x)$ during training of the autoencoders $AE_{\theta}^{\text{KL,T}}$ for the CIFAR-10 models. We find that the optimal value of $T$ for the classifiers on CIFAR-10 equals $0.5$ for the strong attacks. \Cref{tb:table_cifar10_weak} and \Cref{tb:table_cifar10_strong} show that temperature scaling improves the classifier accuracy on the adversarial instances generated by strong attacks like C\&W and SLIDE by an additional $0.93$\% to $1.36$\% for the CIFAR-10 classifiers. Decreasing the temperature too much leads to increasing overconfidence on potentially incorrect predictions since the true labels are not used during training of the autoencoder.

\Cref{tb:table_cifar10_weak} and \Cref{tb:table_cifar10_strong} also illustrate that including the hidden layer divergence using \Cref{eq:loss_hl} for the CIFAR-10 classifiers leads to an accuracy improvement between $1$\% and $2$\% on the C\&W and SLIDE attacks compared to our basic defence mechanism. The feature maps are extracted after the max-pooling layer for the simple CIFAR-10 model and before the activation function in the last residual block of the ResNet-56. As shown in the appendix, the improvement is robust with respect to the choice of extracted hidden layer.

\subsection{White-Box Attacks}

In the case of white-box attacks, C\&W and SLIDE are able to reduce the accuracy of the predictions $C(AE_{\theta}^{\text{KL}}(x_{\text{adv}}))$ on all datasets to almost $0$\%. This does not mean that the attack goes unnoticed or cannot be corrected in practice. 

Moreover, while $x_{\text{adv}}$ manages to bypass the correction mechanism it has a limited impact on the model accuracy itself. For the ResNet-56 classifier on CIFAR-10, the prediction accuracy of $C(x)$ equals $93.15$\% on the original test set. The accuracy of the model predictions on the adversarial instances $C(x_{\text{adv}})$ only drops to respectively $90.57$\% and $92.26$\% after C\&W and SLIDE attacks. 

Importantly, the adversarial attacks are not very transferable. Assume that the autoencoder model under attack has been trained for $N$ epochs. Swapping it for the same model but only trained for $n<N$ epochs drastically reduces the effectiveness of the attack and brings the classification accuracy of $C(AE_{\theta}^{\text{KL}}(x_{\text{adv}}))$ back up from almost $0$\% to respectively $67.59$\% and $73.87$\% for C\&W and SLIDE. In practice this means that we can cheaply boost the strength of the defence by ensembling different checkpoints during the training of $AE_{\theta}^{\text{KL}}$. When the adversarial score $S_{\text{adv}}(x)$ is above a threshold value, the prediction on $x$ is found by applying a weighted majority vote over a combination of the ensemble and the classifier:

\begin{equation}\label{eq:majority_vote}
\begin{split}
    y = \argmax_i &(\sum_{j} w_j 1(h_{\text{AE},j}(x)=i) + \\ &(1 - \sum_{j} w_j) 1(C(x)=i))
\end{split}
\end{equation}

with

\begin{equation}\label{eq:ae_pred_ensemble}
    h_{\text{AE},j}= C(AE_{\theta_j}^{D_{\text{KL}}}(x)).
\end{equation}

By combining autoencoders with different architectures we can further improve the diversification and performance of the ensemble. In order to guarantee the success of the white-box attack, it needs to find a perturbation $\delta$ that fools the weighted majority of the diverse defences and the classifier. MagNet \cite{meng2017magnet} also uses a collection of autoencoders, but trained with the usual MSE loss.

The detector is still effective at flagging the adversarial instances generated by the white-box attacks. This is evidenced in \Cref{fig:fig_roc_wb} by the robust AUC values consistently above $0.97$ for MNIST on all attacks. The AUC for Fashion-MNIST is equal to respectively $0.8544$ and $0.8715$ for C\&W and SLIDE and up to $0.9552$ for FGSM. On CIFAR-10, the AUC's for C\&W and SLIDE are $0.7673$ and $0.7268$, and close to $0.85$ for FGSM for different values of $\epsilon$.

\subsection{Data Drift Detection}\label{sec:drift}

\begin{figure}[t]
\vskip 0.2in
\begin{center}
\centerline{\includegraphics[width=\columnwidth]{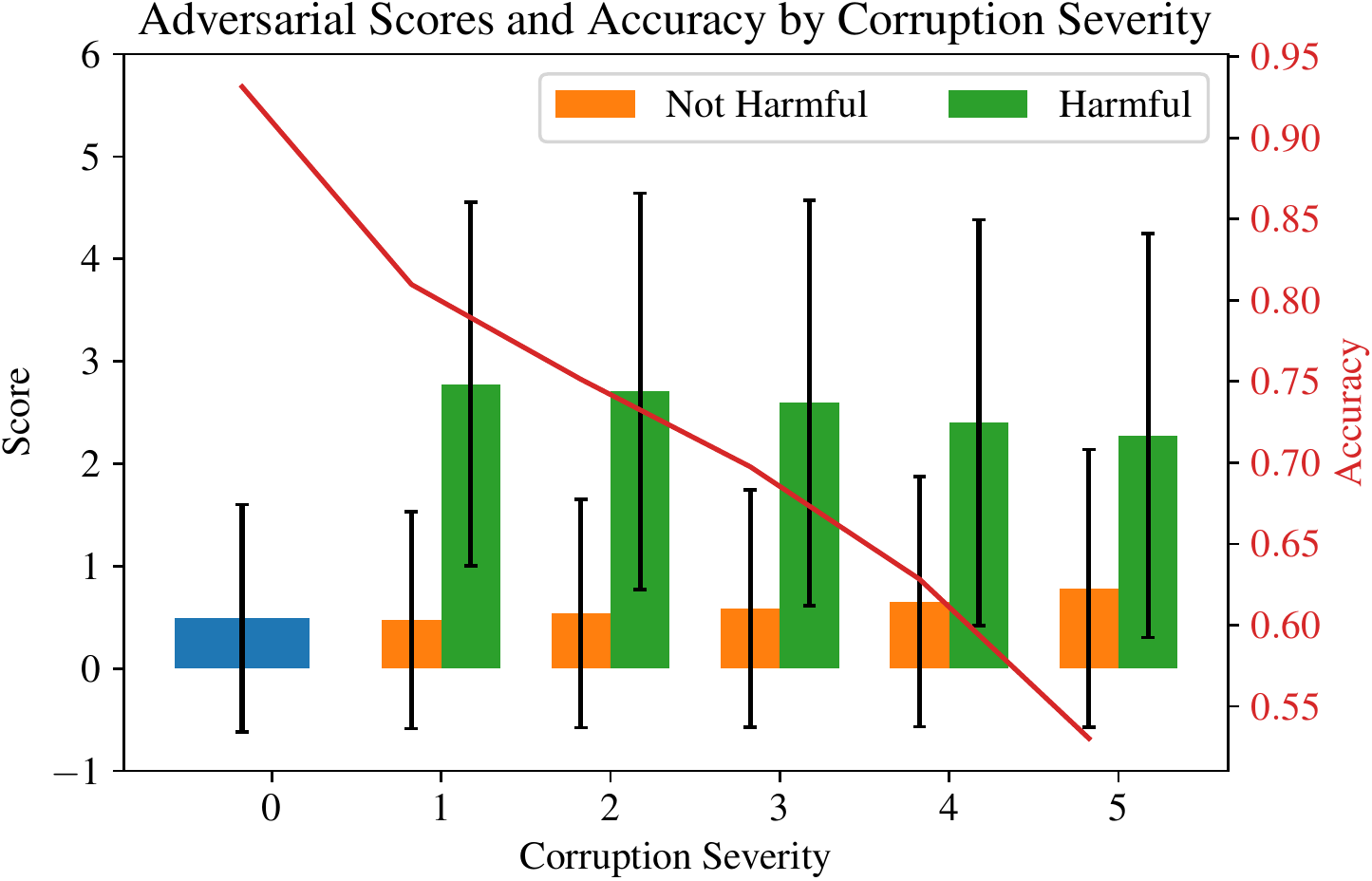}}
\caption{Mean adversarial scores with standard deviations (lhs) and ResNet-56 accuracies (rhs) for increasing data corruption severity levels on CIFAR-10-C. Level 0 corresponds to the original CIFAR-10 test set. Harmful scores are scores from instances which have been flipped from the correct to an incorrect prediction because of the corruption. Not harmful means that the prediction was unchanged after the corruption.}
\label{fig:fig_bar_cd}
\end{center}
\vskip -0.2in
\end{figure}

It is important that safety-critical applications do not suffer from common data corruptions and perturbations. Detection of subtle input changes which reduce the model accuracy is therefore crucial. \citet{NIPS2019_8420} discuss several methods to identify distribution shift and highlight the importance to quantify the harmfulness of the shift. The adversarial detector proves to be very flexible and can be used to measure the harmfulness of the data drift on the classifier. We evaluate the detector on the CIFAR-10-C dataset \cite{HendrycksD19}. The instances in CIFAR-10-C have been corrupted and perturbed by various types of noise, blur, brightness etc. at different levels of severity, leading to a gradual decline in model performance.

\Cref{fig:fig_bar_cd} visualises the adversarial scores $S_{\text{adv}}$ at different levels of corruption severity for the ResNet-56 classifier on CIFAR-10-C compared to CIFAR-10. The average scores for the instances where the predicted class was changed from the correct to an incorrect class due to the data corruption is between $2.91$x and $5.87$x higher than the scores for the instances where the class predictions were not affected by the perturbations. The average $S_{\text{adv}}$ of the negatively affected instances declines slightly with increasing severity because the changes $\delta$ to $x$ are stronger and $D_\text{KL}(M(x + \delta) || M(AE_{\theta}^{\text{KL}}(x + \delta)))$ is not as large. For each level of corruption severity, a two-sided Kolmogorov-Smirnov two sample test \cite{smirnov1939estimate} rejects the null hypothesis that the negatively affected instances are drawn from the same distribution as the samples unaffected by the data corruption with a p-value of $0.0$. As a result, the drift detector provides a robust measure for the harmfulness of the distribution shift.

\section{Conclusions}\label{sec:conclusions}
We introduced a novel method for adversarial detection and correction based on an autoencoder trained with a custom loss function. The loss function aims to match the prediction probability distributions between the original data and the reconstructed instance by the autoencoder. We validate our approach on a variety of grey-box and white-box attacks on the MNIST, Fashion-MNIST and CIFAR-10 datasets. The defence mechanism is very effective at detecting and correcting strong grey-box attacks like Carlini-Wagner or SLIDE and remains efficient for white-box attack detection. Interestingly, the white-box attacks are not very transferable between different autoencoders or even between the defence mechanism and the standalone classification model. This is a promising area for future research and opens opportunities to build robust adversarial defence systems. The method is also successful in detecting common data corruptions and perturbations which harm the classifier's performance. We illustrate the effectiveness of the method on the CIFAR-10-C dataset. To facilitate the practical use of the adversarial detection and correction system we provide an open source library with our implementation of the method \cite{alibi-detect}.

\section*{Acknowledgements}

The authors would like to thank Janis Klaise and Alexandru Coca for fruitful discussions on adversarial detection, and Seldon Technologies Ltd. for providing the time and computational resources to complete the project.

\bibliography{references}
\bibliographystyle{icml2020}

\appendix
\section{Models}

All the models are trained on a NVIDIA GeForce RTX 2080 GPU.

\subsection{MNIST}

The classification model consists of 2 convolutional layers with respectively 64 and 32 $2\times 2$ filters and $\text{ReLU}$ activations. Each convolutional layer is followed by a $2\times 2$ max-pooling layer and a dropout with fraction 30\%. The output of the second pooling layer is flattened and fed into a fully connected layer of size 256 with $\text{ReLU}$ activation and 50\% dropout. This dense layer is followed by a softmax output layer over the 10 classes. The model is trained with an Adam optimizer for 20 epochs with batch size 128 and learning rate 0.001 on MNIST images scaled to $[0, 1]$ and reaches a test accuracy of 99.28\%.

The autoencoder for MNIST has 3 convolutional layers in the encoder with respectively 64, 128 and 512 $4\times 4$ filters with stride 2, $\text{ReLU}$ activations and zero padding. The output of the last convolution layer in the encoder is flattened and fed into a linear layer which outputs a 10-dimensional latent vector. The decoder takes this latent vector, feeds it into a linear layer with $\text{ReLU}$ activation and output size of 1568. This output is reshaped and passed through 3 transposed convolution layers with 64, 32 and 1 $3\times 3$ filters and zero padding. The first 2 layers have stride 2 while the last layer has a stride of 1. The autoencoder is trained with the different custom loss terms for 50 epochs using an Adam optimizer with batch size 128 and learning rate 0.001.

\subsection{Fashion-MNIST}

The classification model is very similar to the MNIST classifier. It consists of 2 blocks of convolutional layers. Each block has 2 convolutional layers with $\text{ReLU}$ activations and zero padding followed by a $2\times 2$ max-pooling layer and dropout with fraction 30\%. The convolutions in the first and second block have respectively 64 and 32 $2\times 2$ filters. The output of the second block is flattened and fed into a fully connected layer of size 256 with $\text{ReLU}$ activation and 50\% dropout. This dense layer is followed by a softmax output layer over the 10 classes. The model is trained with an Adam optimizer for 40 epochs with batch size 128 on Fashion-MNIST images scaled to $[0, 1]$ and reaches a test accuracy of 93.62\%.

The autoencoder architecture and training procedure is exactly the same as the one used for the MNIST dataset.

\subsection{CIFAR-10}

We train 2 different classification models on CIFAR-10: a simple network with test set accuracy of 80.24\% and a ResNet-56\footnote{https://github.com/tensorflow/models} with an accuracy of 93.15\%. The simple classifier has the same architecture as the Fashion-MNIST model and is trained for 300 epochs. The ResNet-56 is trained with an SGD optimizer with momentum 0.9 for 300 epochs with batch size 128. The initial learning rate is 0.01, which is decreased with a factor of 10 after 91, 136 and 182 epochs. The CIFAR-10 images are standardised on an image-by-image basis for each model.

The autoencoder for CIFAR-10 has 3 convolutional layers in the encoder with respectively 32, 64 and 256 $4\times 4$ filters with stride 2, $\text{ReLU}$ activations, zero padding and $L1$ regularisation. The output of the last convolution layer in the encoder is flattened and fed into a linear layer which outputs a 40-dimensional latent vector. The decoder takes this latent vector, feeds it into a linear layer with $\text{ReLU}$ activation and output size of 2048. This output is reshaped and passed through 3 transposed convolution layers with 256, 64 and 3 $4\times 4$ filters with stride 2, zero padding and $L1$ regularisation. The autoencoder is trained with the different custom loss terms for 50 epochs using an Adam optimizer with batch size 128 and learning rate 0.001.

The adversarial detection mechanism is also tested with an autoencoder where the $32\times 32 \times 3$ input is flattened, fed into a dense layer with ReLU activation, $L1$ regularisation and output size 512 before being projected by a linear layer on the 40-dimensional latent space. The decoder consists of one hidden dense layer with ReLU activation, $L1$ regularisation and output size 512 and a linear output layer which projects the data back to the $32\times 32 \times 3$ feature space after reshaping. Again, the  autoencoder is trained with the different custom loss terms for 50 epochs using an Adam optimizer with batch size 128 and learning rate 0.001.

\section{Attacks}
\subsection{Carlini-Wagner (C\&W)}

\begin{figure}[!h]
\vskip 0.2in
\begin{center}
\centerline{\includegraphics[width=\columnwidth]{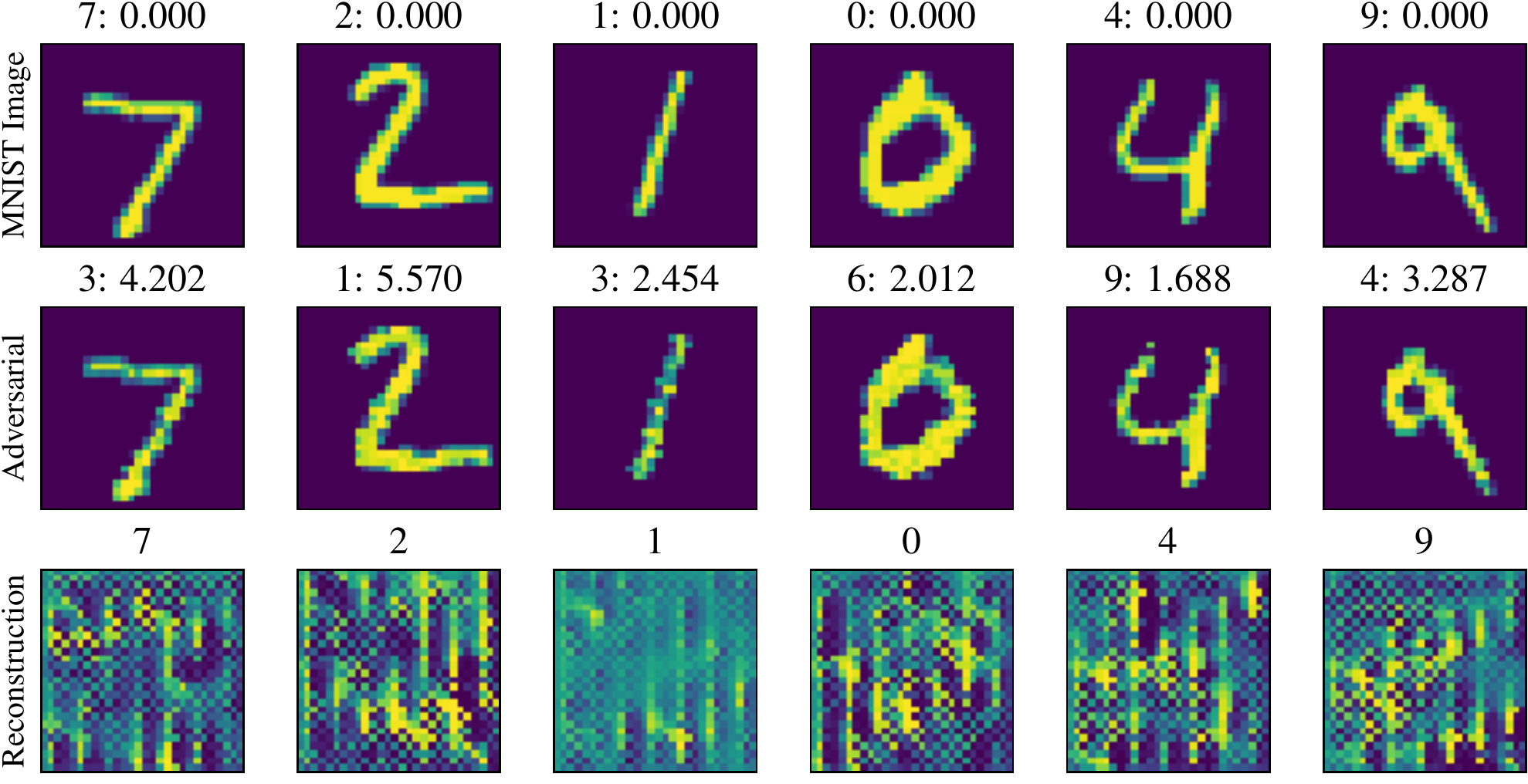}}
\caption{C\&W attack on MNIST. The rows illustrate respectively the original, adversarial and reconstructed instance with their model predictions and adversarial scores.}
\label{fig:fig_mnist_cw_sm}
\end{center}
\vskip -0.2in
\end{figure}

\begin{figure}[!h]
\vskip 0.2in
\begin{center}
\centerline{\includegraphics[width=\columnwidth]{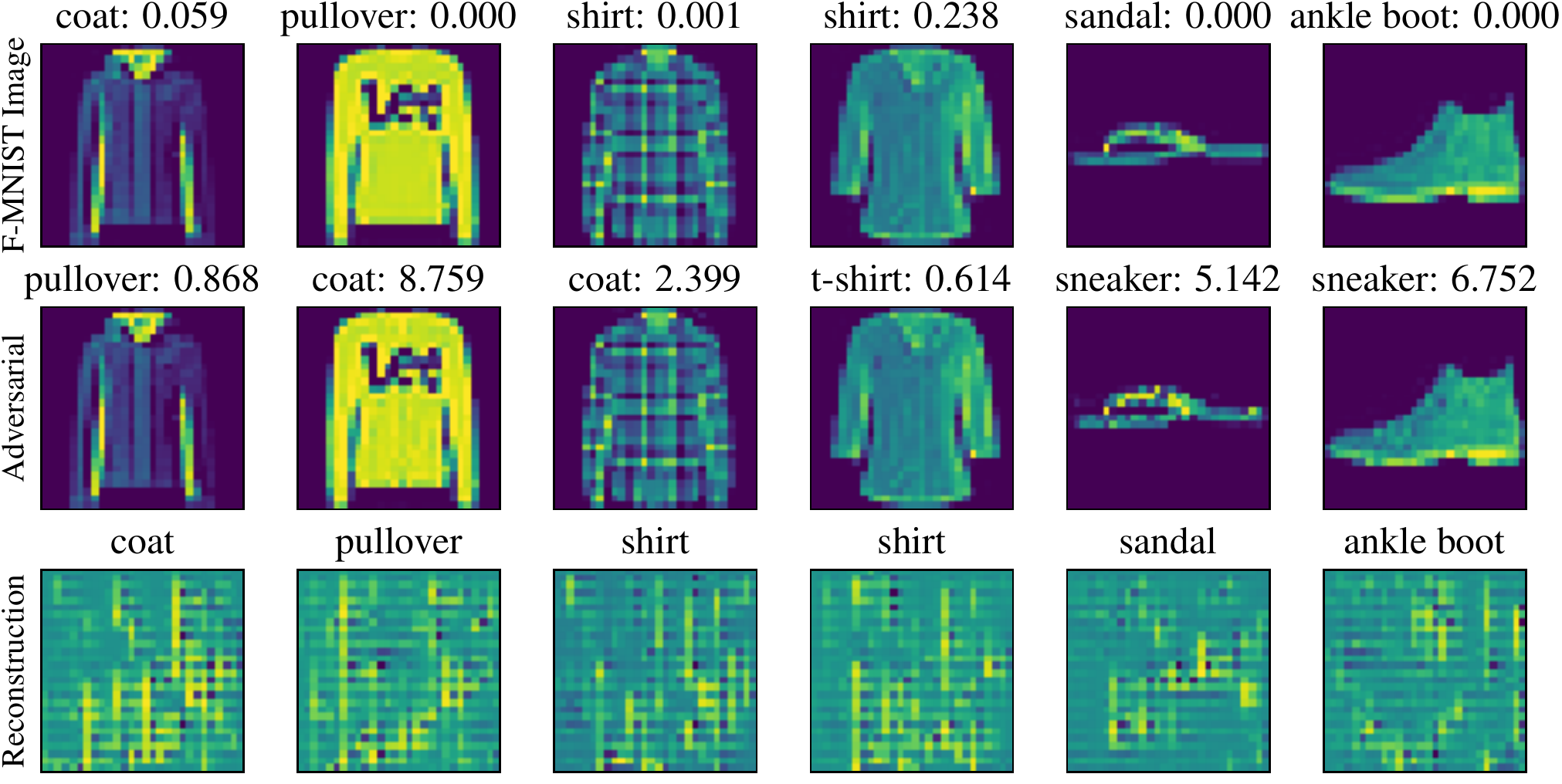}}
\caption{C\&W attack on Fashion-MNIST. The rows illustrate respectively the original, adversarial and reconstructed instance with their model predictions and adversarial scores.}
\label{fig:fig_fashion_mnist_cw_sm}
\end{center}
\vskip -0.2in
\end{figure}

\begin{figure}[!h]
\vskip 0.2in
\begin{center}
\centerline{\includegraphics[width=\columnwidth]{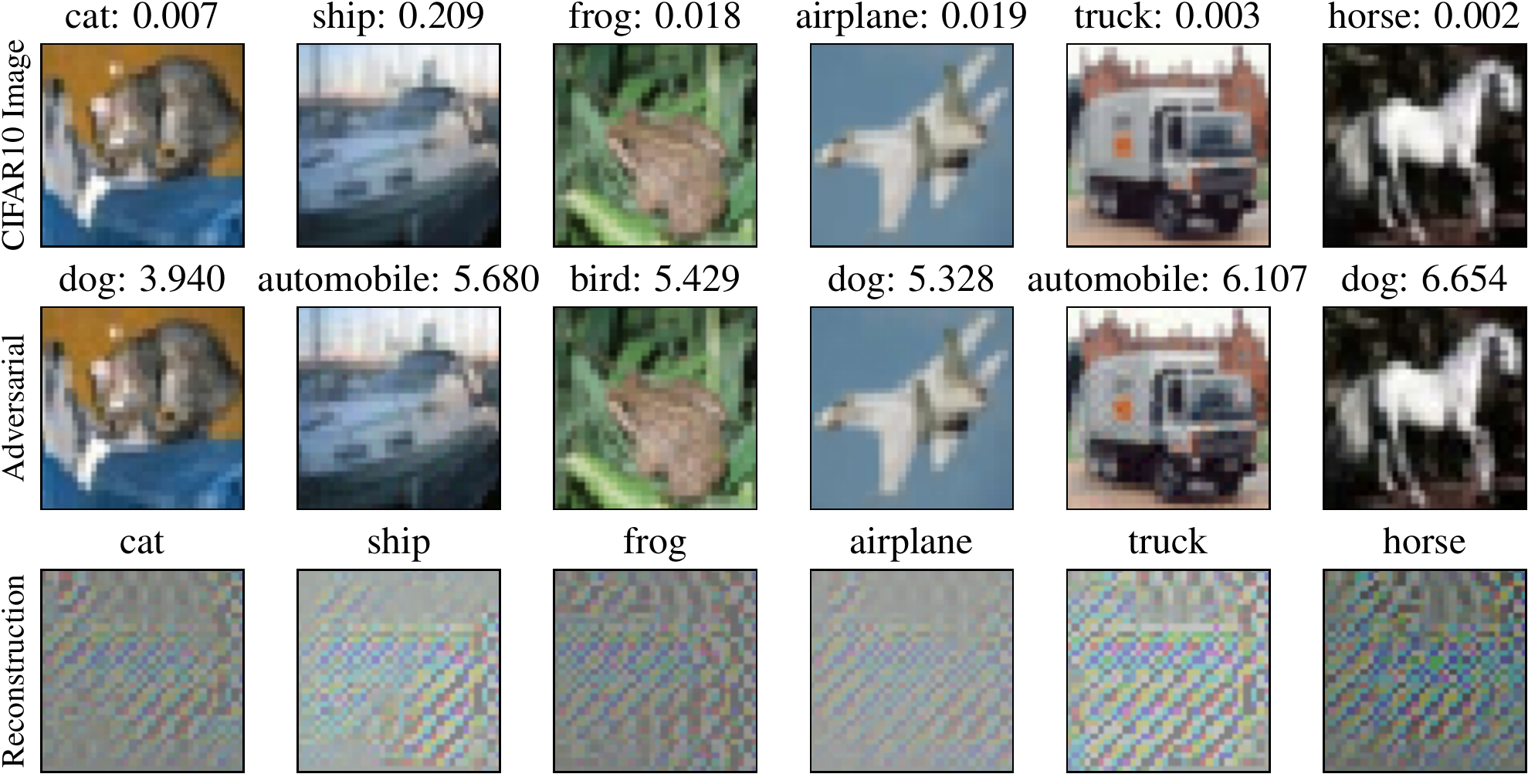}}
\caption{C\&W attack on CIFAR-10 using the ResNet-56 model. The rows illustrate respectively the original, adversarial and reconstructed instance with their model predictions and adversarial scores.}
\label{fig:fig_cifar10_cw_sm}
\end{center}
\vskip -0.2in
\end{figure}

On the MNIST dataset, the initial constant $c$ is equal to 100. 7 binary search steps are applied to update $c$ to a more suitable value. The maximum number of iterations for the attack for each $c$ is 200 and the learning rate of the Adam optimizer used during the iterations equals 0.1. Except for the number of binary search steps which is increased to 9, the hyperparameters for Fashion-MNIST are the same as for MNIST. For CIFAR-10, the initial constant $c$ is set at 1 with 9 binary search steps to find the optimal value. The learning rate and maximum number of iterations are decreased to respectively 0.01 and 100. \Cref{fig:fig_mnist_cw_sm}, \Cref{fig:fig_fashion_mnist_cw_sm} and \Cref{fig:fig_cifar10_cw_sm} illustrate a number of examples for the C\&W attack on each dataset. The first row shows the original instance with the correct model prediction and adversarial score. The second row illustrates the adversarial example with the adversarial score and incorrect prediction. The last row shows the reconstruction by the adversarial detector with the corrected model prediction.

\subsection{SLIDE}

\begin{figure}[!h]
\vskip 0.2in
\begin{center}
\centerline{\includegraphics[width=\columnwidth]{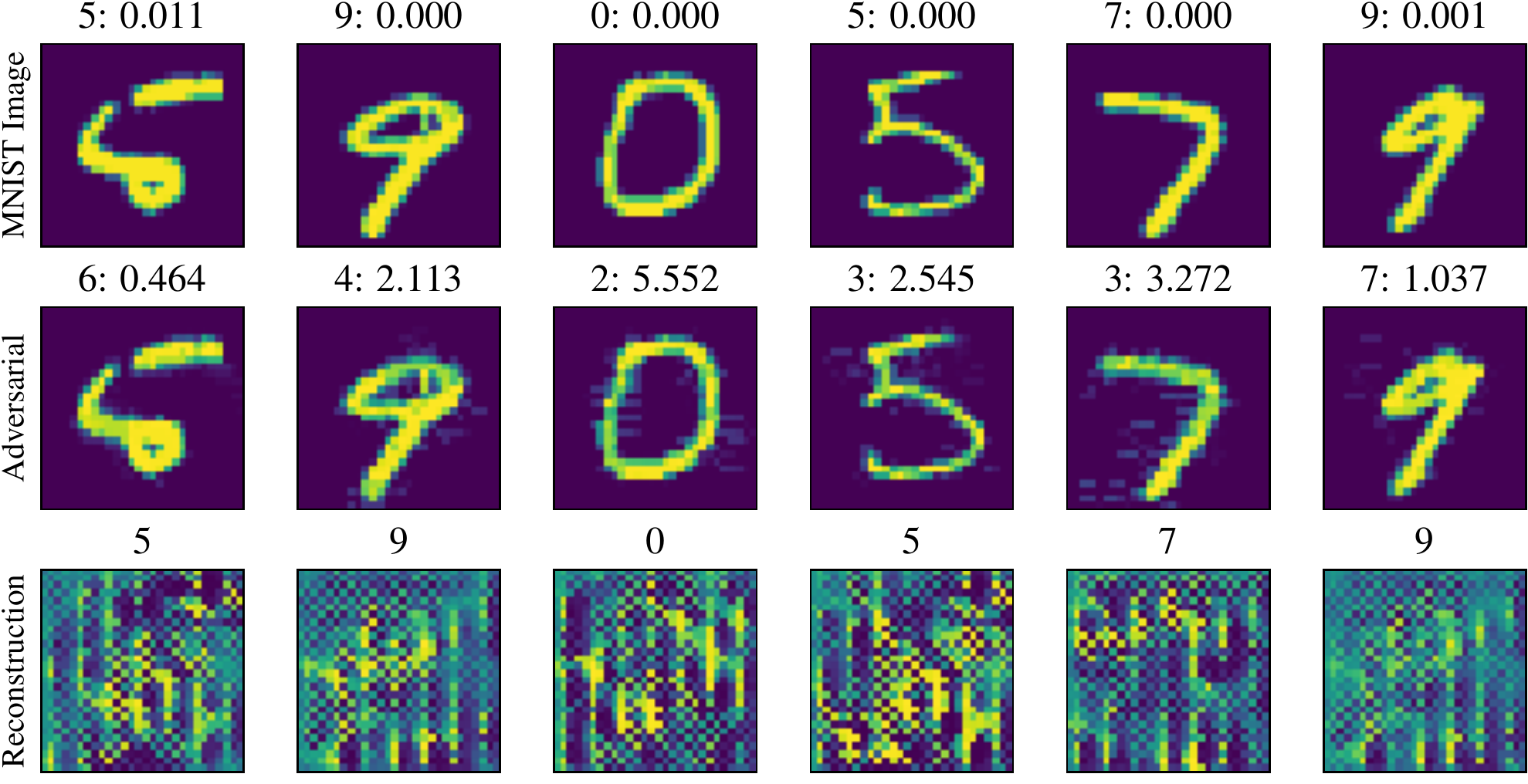}}
\caption{SLIDE attack on MNIST. The rows illustrate respectively the original, adversarial and reconstructed instance with their model predictions and adversarial scores.}
\label{fig:fig_mnist_sl1bia_sm}
\end{center}
\vskip -0.2in
\end{figure}

\begin{figure}[!h]
\vskip 0.2in
\begin{center}
\centerline{\includegraphics[width=\columnwidth]{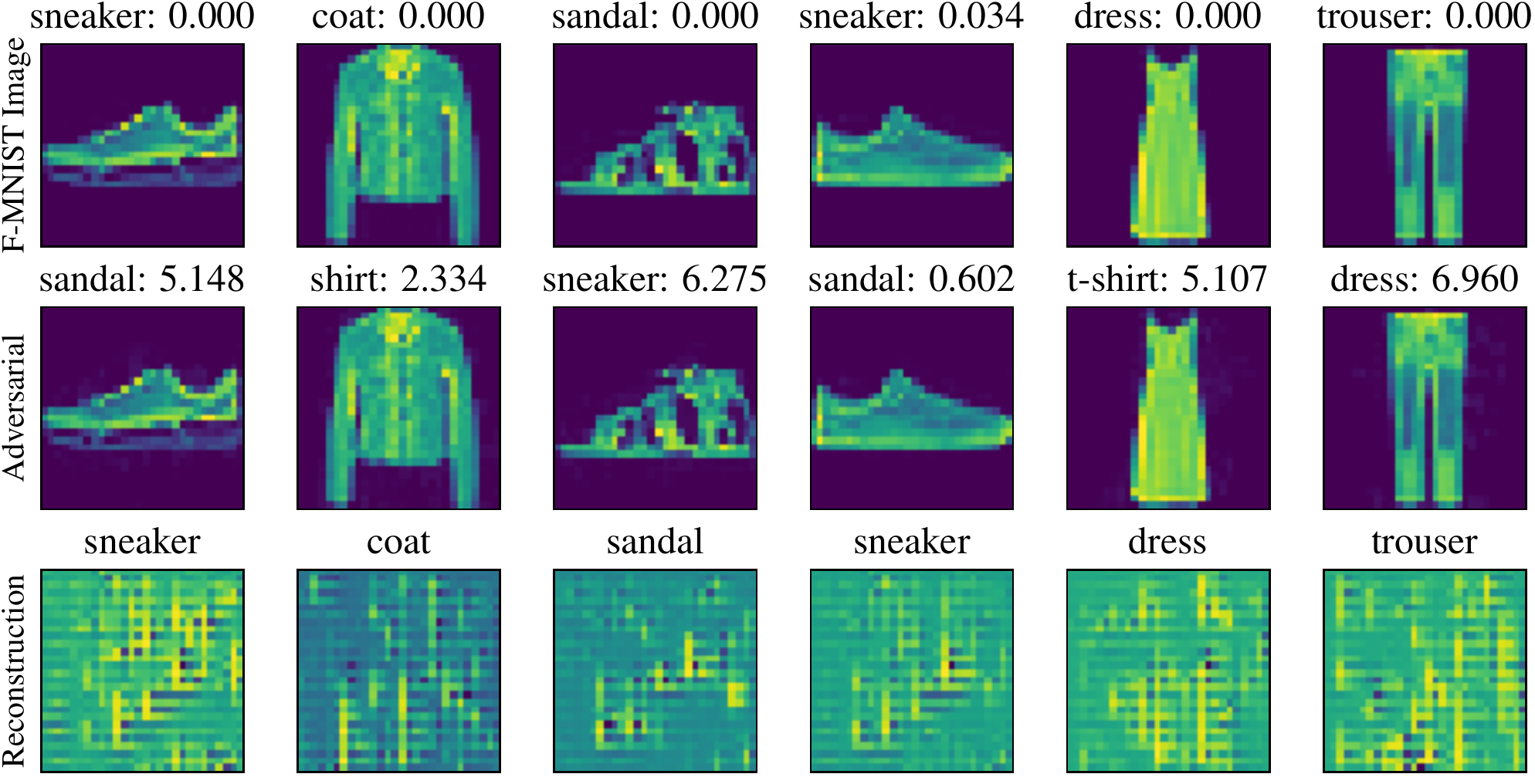}}
\caption{SLIDE attack on Fashion-MNIST. The rows illustrate respectively the original, adversarial and reconstructed instance with their model predictions and adversarial scores.}
\label{fig:fig_fashion_mnist_sl1bia_sm}
\end{center}
\vskip -0.2in
\end{figure}

\begin{figure}[!h]
\vskip 0.2in
\begin{center}
\centerline{\includegraphics[width=\columnwidth]{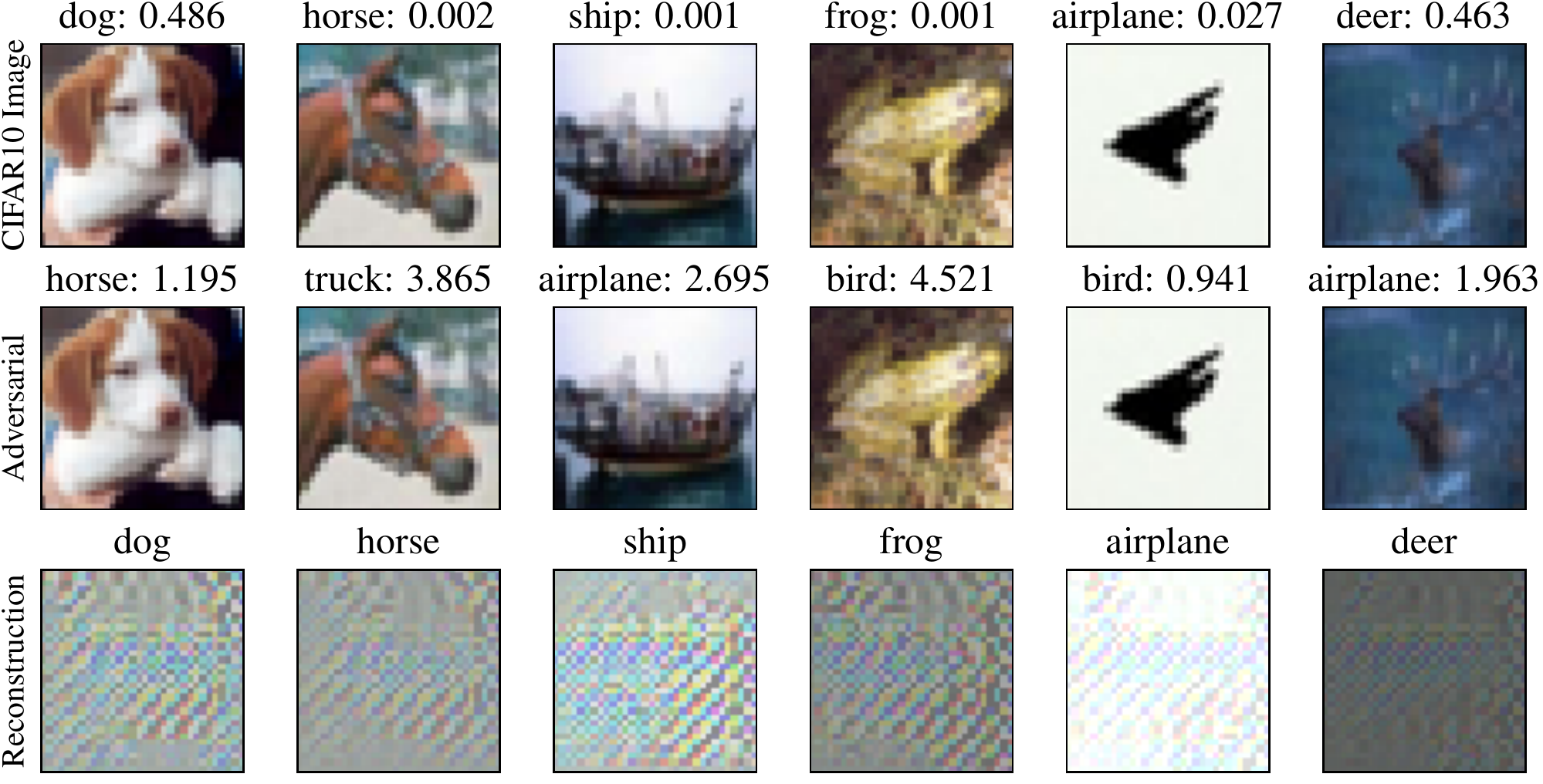}}
\caption{SLIDE attack on CIFAR-10 using the ResNet-56 model. The rows illustrate respectively the original, adversarial and reconstructed instance with their model predictions and adversarial scores.}
\label{fig:fig_cifar10_sl1bia_sm}
\end{center}
\vskip -0.2in
\end{figure}

The hyperparameters of the attack remain unchanged for the different datasets. The percentile $q$ is equal to 80, the $\ell_{1}$-bound $\epsilon$ is set at 0.1, the step size $\gamma$ equals 0.05 and the number of steps $k$ equals 10. \Cref{fig:fig_mnist_sl1bia_sm}, \Cref{fig:fig_fashion_mnist_sl1bia_sm} and \Cref{fig:fig_cifar10_sl1bia_sm} show a number of examples for the SLIDE attack on each dataset. Similar to C\&W, the rows illustrate respectively the original, adversarial and reconstructed instance with their model predictions and adversarial scores.

\subsection{Fast Gradient Sign Method (FGSM)}

\begin{figure}[!t]
\vskip 0.2in
\begin{center}
\centerline{\includegraphics[width=\columnwidth]{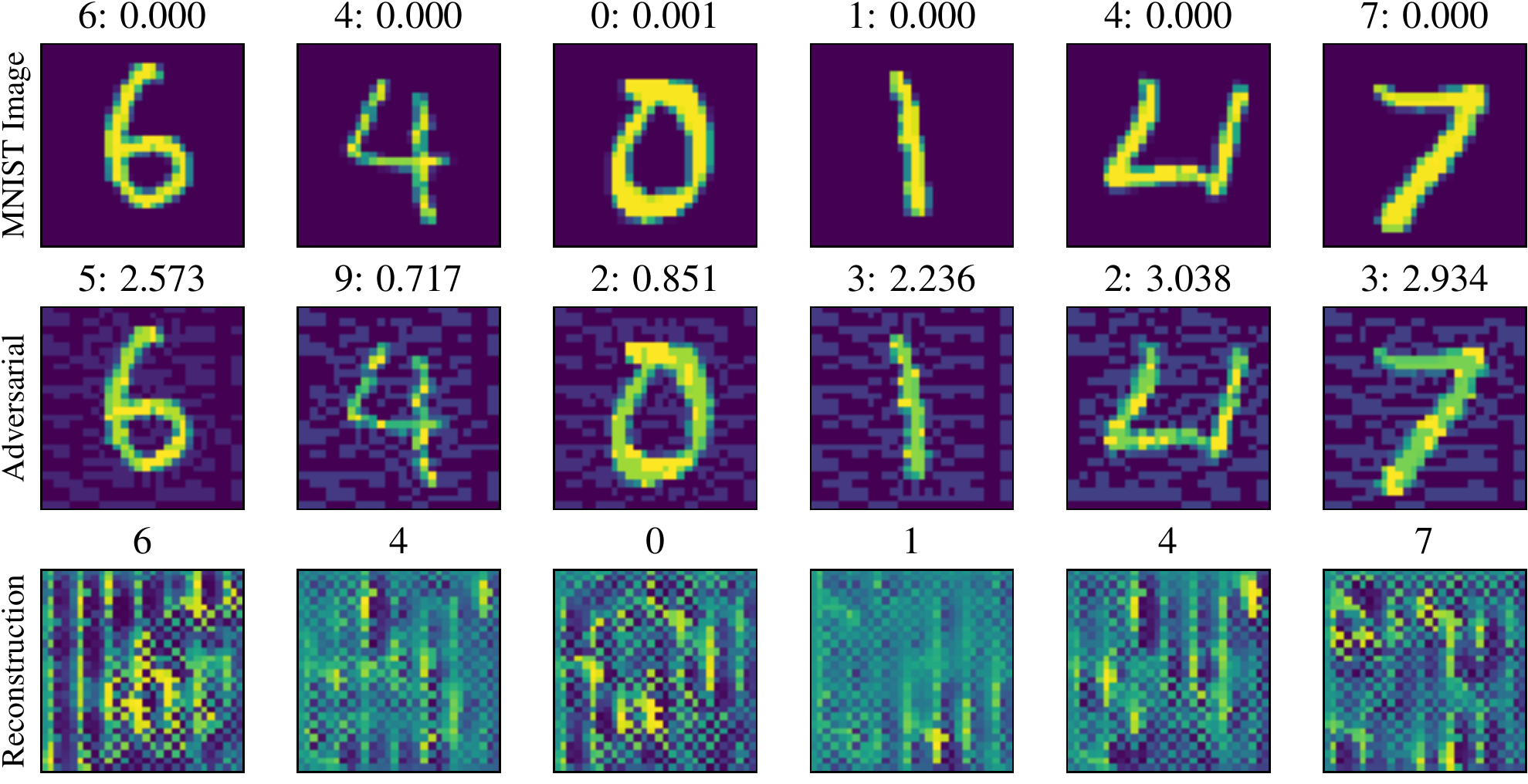}}
\caption{FGSM attack with $\epsilon$ 0.2 on MNIST. The rows illustrate respectively the original, adversarial and reconstructed instance with their model predictions and adversarial scores.}
\label{fig:fig_mnist_fgsm_2_sm}
\end{center}
\vskip -0.2in
\end{figure}

\begin{figure}[!t]
\vskip 0.2in
\begin{center}
\centerline{\includegraphics[width=\columnwidth]{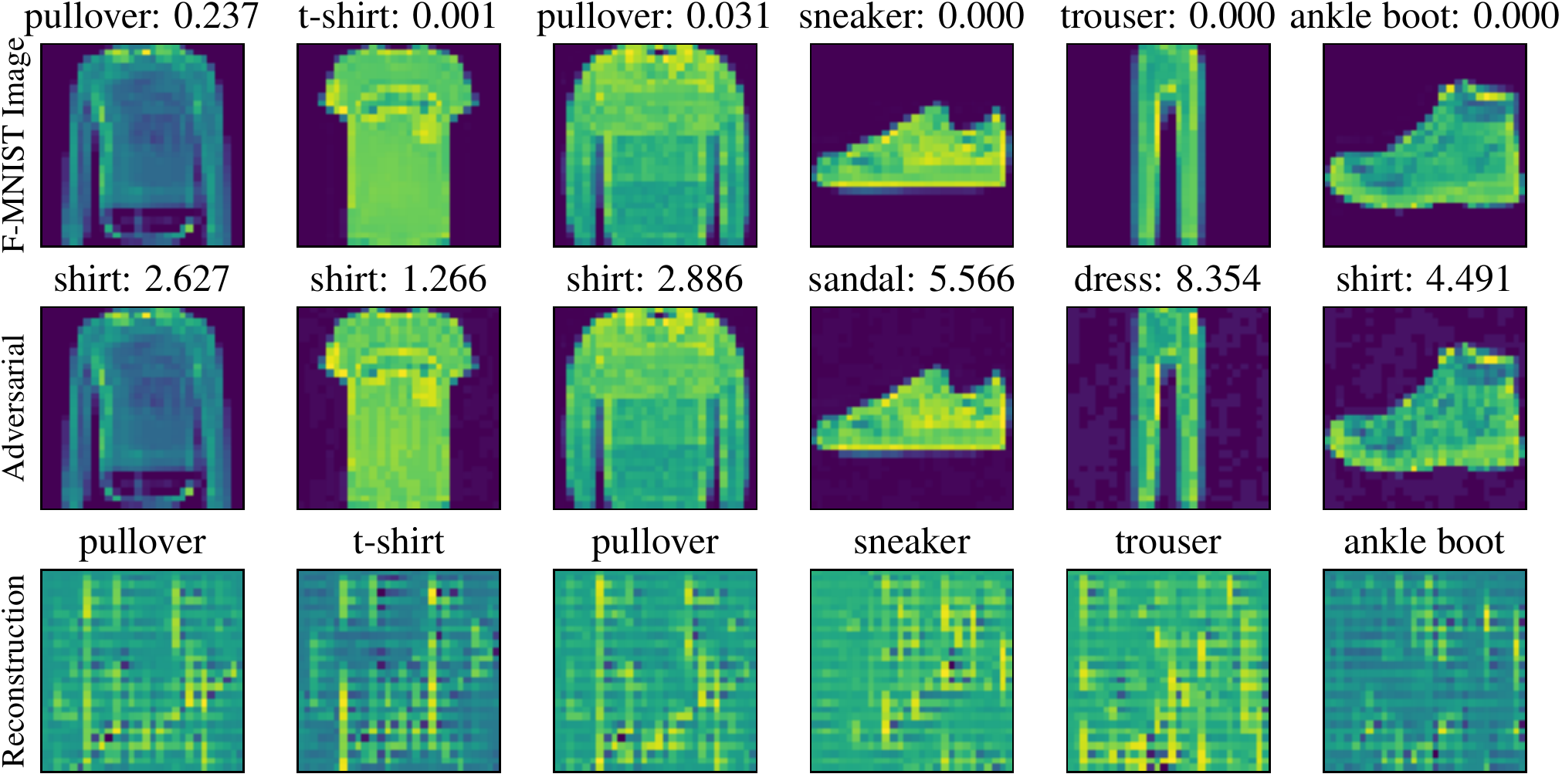}}
\caption{FGSM attack with $\epsilon$ 0.1 on Fashion-MNIST. The rows illustrate respectively the original, adversarial and reconstructed instance with their model predictions and adversarial scores.}
\label{fig:fig_fashion_mnist_fgsm_1_sm}
\end{center}
\vskip -0.2in
\end{figure}

\begin{figure}[!t]
\vskip 0.2in
\begin{center}
\centerline{\includegraphics[width=\columnwidth]{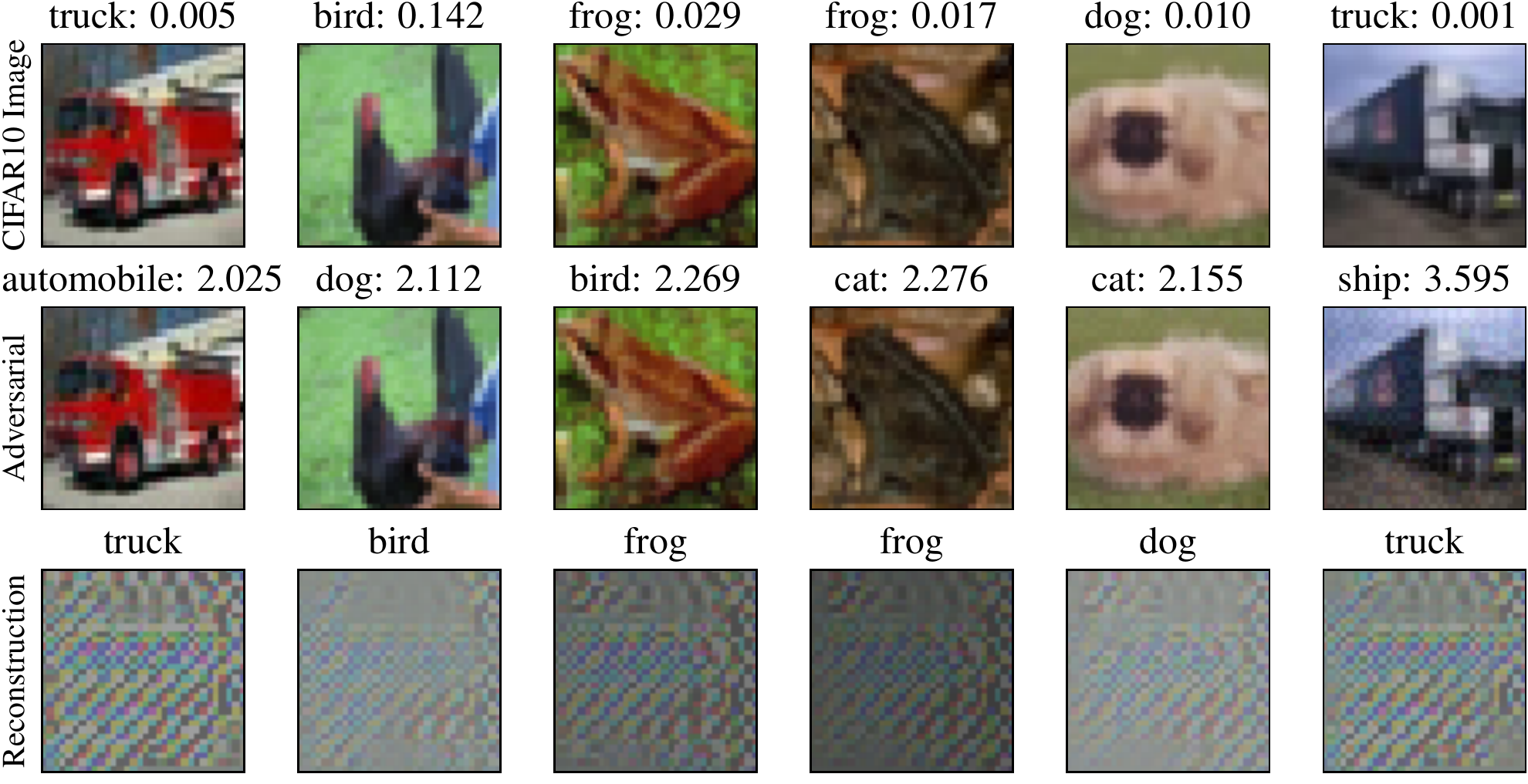}}
\caption{FGSM attack with $\epsilon$ 0.1 on CIFAR-10 using the ResNet-56 model. The rows illustrate respectively the original, adversarial and reconstructed instance with their model predictions and adversarial scores.}
\label{fig:fig_cifar10_fgsm_1_sm}
\end{center}
\vskip -0.2in
\end{figure}

$\epsilon$ values of 0.1, 0.2 and 0.3 are used for the FGSM attack on each dataset. The attacks last for 1000 iterations. \Cref{fig:fig_mnist_fgsm_2_sm}, \Cref{fig:fig_fashion_mnist_fgsm_1_sm} and \Cref{fig:fig_cifar10_fgsm_1_sm} show a number of examples for the FGSM attack on each dataset. Similar to C\&W, the rows illustrate respectively the original, adversarial and reconstructed instance with their model predictions and adversarial scores.

\section{Hidden Layer K-L Divergence}

\Cref{tb:table_cifar10_weak_hkld} and \Cref{tb:table_cifar10_strong_hkld} show the robustness of the choice of hidden layer to extract the feature map from before feeding it into a linear layer and applying the softmax function.

\begin{table}[t]
    \centering
    \caption{CIFAR-10 test set accuracy for a simple CNN classifier on both the original and adversarial instances with and without the defence. $AE^{\text{KL}}$ is the defence mechanisms trained with the $D_{\text{KL}}$ loss function. $AE^{\text{KL, HL}}$ extends the methodology to one of the hidden layers. HL1, HL2, HL3 and HL4 refer to respectively the first max-pooling layer, the third convolutional layer, the dropout layer after the second convolution block and the output of the flattening layer in the CNN model. HL1 is projected on a 50-dimensional vector, HL2 and HL3 on 40-dimensional vectors and HL4 on 10 dimensions.}
    \vskip 0.15in
    \resizebox{\columnwidth}{!}{%
    \begin{tabular}{@{}cccccccc@{}}
    \toprule
    \textbf{Attack} & \textbf{No Attack} & \textbf{No Defence} & $\mathbf{AE^{\text{KL}}}$ & $\mathbf{AE^{\text{KL, HL1}}}$  & $\mathbf{AE^{\text{KL, HL2}}}$ & $\mathbf{AE^{\text{KL, HL3}}}$ & $\mathbf{AE^{\text{KL, HL4}}}$ \\ \midrule
    CW & $0.8024$ & $0.0001$ & $0.7551$ & $0.7662$ & $0.7687$ & $0.7655$ & $0.7688$ \\
    SLIDE & $0.8024$ & $0.0208$ & $0.7704$ & $0.7807$ & $0.7835$ & $0.7838$ & $0.7864$ \\ \bottomrule
    \end{tabular}%
    }
    \label{tb:table_cifar10_weak_hkld}
\end{table}

\begin{table}[t]
    \centering
    \caption{CIFAR-10 test set accuracy for a ResNet-56 classifier on both the original and adversarial instances with and without the defence. $AE^{\text{KL}}$ is the defence mechanisms trained with the $D_{\text{KL}}$ loss function. $AE^{\text{KL, HL}}$ extends the methodology to one of the hidden layers. HL1, HL2, HL3 and HL4 refer to respectively hidden layers 140, 160, 180 and 200 in the ResNet-56 model. HL1 to HL4 are all projected on 20-dimensional vectors.}
    \vskip 0.15in
    \resizebox{\columnwidth}{!}{%
    \begin{tabular}{@{}cccccccc@{}}
    \toprule
    \textbf{Attack} & \textbf{No Attack} & \textbf{No Defence} & $\mathbf{AE^{\text{KL}}}$ & $\mathbf{AE^{\text{KL, HL1}}}$  & $\mathbf{AE^{\text{KL, HL2}}}$ & $\mathbf{AE^{\text{KL, HL3}}}$ & $\mathbf{AE^{\text{KL, HL4}}}$ \\ \midrule
    CW & $0.9315$ & $0.0000$ & $0.8048$ & $0.8049$ & $0.8094$ & $0.8152$ & $0.8153$ \\
    SLIDE & $0.9315$ & $0.0000$ & $0.8159$ & $0.8251$ & $0.8277$ & $0.8347$ & $0.8360$ \\ \bottomrule
    \end{tabular}%
    }
    \label{tb:table_cifar10_strong_hkld}
\end{table}

\end{document}